\documentclass{optica-article}

\journal{optcon}

\articletype{Research Article}

\usepackage{lineno}
\usepackage{color}
\usepackage{graphicx}

\usepackage{subcaption}
\usepackage{cleveref}  
\usepackage{xcolor}
\usepackage{array}

\begin{document}

\title{Neural Invertible Variable-degree Optical Aberrations Correction}

\author{Shuang Cui\authormark{1,2}, Bingnan Wang\authormark{1,2}, Quan Zheng\authormark{1,*}}

\address{\authormark{1}Institute of Software, Chinese Academy of Sciences, Beijing 100190, China\\
	\authormark{2}University of Chinese Academy of Sciences, Beijing 100049, China\\
}

\email{\authormark{*}zhengquan@iscas.ac.cn}



\begin{abstract}
Optical aberrations of optical systems cause significant degradation of imaging quality. Aberration correction by sophisticated lens designs and special glass materials generally incurs high cost of manufacturing and the increase in the weight of optical systems, thus recent work has shifted to aberration correction with deep learning-based post-processing. Though real-world optical aberrations vary in degree, existing methods cannot eliminate variable-degree aberrations well, especially for the severe degrees of degradation. Also, previous methods use a single feed-forward neural network and suffer from information loss in the output. To address the issues, we propose a novel aberration correction method with an invertible architecture by leveraging its information-lossless property. Within the architecture, we develop conditional invertible blocks to allow the processing of aberrations with variable degrees. Our method is evaluated on both a synthetic dataset from physics-based imaging simulation and a real captured dataset. Quantitative and qualitative experimental results demonstrate that our method outperforms compared methods in correcting variable-degree optical aberrations.
\end{abstract}

\section{Introduction}
Optical aberrations are defects introduced in the design, manufacturing and assembly of camera lenses. The defects cause the incident light to diffuse and fail to focus to form a sharp image, producing images with blurry and dispersive appearance\cite{schuler2011non}. Despite the rapid development of optical design and digital imaging technologies, they still can not totally avoid image degradation caused by optical aberrations, especially from lightweight cheap lenses and camera lenses of smart phones, whose optical aberrations are relatively significant.

In practice, optical aberrations are mitigated by incorporating an array of lenses, using aspheric lenses, and employing special glass materials for lenses. However, the increase in lens types and lens materials poses challenges to the manufacturing process and raises production costs. Therefore, recent works have shifted the genre of aberration corrections from sophisticated lens design to post-processing \cite{schuler2012blind}. 

Currently, two main technical routes have been proposed for optical aberrations correction in post-processing. One is the model-driven traditional methods, which assume an image degradation model, use various natural image priors, optimize in multiple iterations to find the degradation kernel, and perform deconvolution to obtain the sharp image. However, traditional methods are not robust enough in dealing with spatially varying degradation. The other is the data-driven deep learning methods, which have recently become increasingly popular. These methods utilize training data to train the neural networks and recover sharp images from degraded images. However, most of them can not deal with variable-degree aberrations since they are exclusively designed for a specific degree of degradation. In addition, they leverage a single feed-forward autoencoder architecture and usually suffer from information loss during the encoding and decoding process.

To address the information loss problem, we propose a novel aberration correction method based on invertible neural networks (INNs) to learn the transformation from aberration images to aberration-free images, where the invertible design assures that the neural networks can preserve information, especially the details of images\cite{liu2021invertible,zhang2021ivpf}. Due to the limited nonlinear transformation ability of invertible neural networks \cite{dinh2015nice}, we introduce a feature extraction module to improve the non-linear transformation capability of the INNs. In order to process optical aberrations with variable degrees, we propose enhanced conditional encoding modules, which use the degradation degree of the aberration image as the input. This provides our method with the capability to restore sharp images from input images with variable degradation degrees.

Since images captured by cameras are naturally degraded by optical aberrations, large-scale real paired aberration and aberration-free images are unavailable. To mitigate this issue, we establish an imaging simulation process to synthesize realistic degraded images from sharp reference images. The image simulation process incorporates lens parameters of an optical system and is able to perform physics-based ray tracing to simulate the optical aberrations with variable degrees. We leverage this approach to produce large-scale paired datasets for the proposed aberration correction method.

Experimental results show that the proposed method achieves better numerical metrics and visual effects on both synthetic images and real aberration-degraded images. The visual results verify that our method can recover more details by the inference along the forward direction, thanks to the highly invertible design in the architecture. Meanwhile, our method brings the benefit that it can synthesize aberration images from sharp images along the reverse direction. The contributions of this paper can be summarized as follows: 

\begin{itemize}
\item[$\bullet$] We design an imaging simulation process based on ray tracing and spatial convolution to generate large-scale paired datasets with variable degradation degrees.
\item[$\bullet$] We propose an invertible neural network architecture for optical aberration correction that can largely alleviate the information loss problem and improve image quality.
\item[$\bullet$] We introduce conditional encoding modules for the invertible neural network to deal with varying degrees of optical aberrations.
\end{itemize}

\section{Related work}
\subsection{Optical Aberrations Correction}
Due to the inherent optical aberrations of optical systems, the captured image will be degraded. This degradation can hardly be totally avoided by sophisticated optical system design, so recent works turn to post-processing for removing aberrations. The current methods mostly perform the process in two steps: first, estimate the point spread function (PSF) of the target optical system, and then use the non-blind deconvolution or deep neural networks to restore the image. Optical aberrations are spatially varying, and the methods for obtaining spatially varying PSF can be divided into three categories: real shooting-based methods, calibration-based methods and optical simulation-based methods. In \cite{shih2012image}, the point spread function was directly measured  by imaging the pinhole grid pattern in a dark room. The work in \cite{heide2013high} used the frame random mode to calibrate PSFs. The work in \cite{chen2021optical} calculated the PSFs by raytracing and coherent superposition in a simulation manner.

After obtaining PSFs, some methods use a deconvolution process \cite{li2021universal,schuler2011non} to solve the linear inverse problem. The scholars of \cite{eboli2022fast} used a two-step scheme to correct the aberration of a single image, and then used a convolutional neural network to remove the remaining chromatic aberration in the image. However, deconvolution involves a complex iterative process. Due to the strong fitting ability of the deep neural network, some methods \cite{tian2019dnn,chen2021extreme,chen2021optical,lin2022non} used the autoencoder-based architecture to restore degraded images. The work in \cite{li2021universal} designed a PSF aware neural network, which takes degraded images and PSF images as inputs and generates latent high-quality images by combining depth prior. The work in \cite{lin2022non} proposed a frequency-based adaptive block, which was inserted into the neural network to perform feature based deconvolution to correct non-uniform blur. However, these networks need to be calibrated PSFs as the input. The scholars of \cite{chen2021extreme} proposed an end-to-end neural network to remove the aberrations in the input image. However, their architecture based on a feed-forward autoencoder was unable to deal with varying degrees of aberrations. In contrast, our method does not need a complex PSF estimation process and the proposed a conditional invertible neural network allows to correct degradations with variable degrees.

\subsection{Invertible Neural Networks}
The development of invertible neural networks (INN) can be traced back to nonlinear independent component estimation (NICE), which was proposed by \cite{dinh2015nice}. It learns the nonlinear bi-directional mapping of input data to latent space in an unsupervised way. The forward calculation process and the reverse process have shared model parameters. Based on this, RevNet \cite{gomez2017reversible} was put forward, which can complete the back propagation of the network without storing activation. In this way, the memory consumption of the model can be greatly reduced. In order to deal with image-related tasks, the scholars of \cite{dinh2016density} introduced the convolution layer and multi-scale layer in the coupling model to reduce the computing cost and improve the model regularization ability. The work in \cite{jacobsen2018revnet} built a reversible network architecture i-RevNet  based on RevNet, which retains all information of the input signal in all intermediate representations except the last layer. The article also proves that information loss is not a necessary condition when learning can be generalized to the representation of unfamiliar data. The work in \cite{kingma2018glow} proposed an effective reversible 1 $\times$ 1 convolution block and Glow, which can synthesize and process large images efficiently and realistically. 

Because of the information-lossless and powerful generation ability, INN has been used in many image restoration tasks. The work in \cite{xiao2020invertible} used INN to learn the reversible bijection transformation of image downscaling and upscaling to achieve information-lossless image rescaling. Scholars of \cite{liu2021invertible} designed a reversible neural network for denoising tasks. In the forward process, the noise image is mapped to a low resolution image and a latent representation space; In the reverse process, sampling from a prior distribution will replace the latent representation to discard the noise. Other image restoration tasks using INN include image decolorization \cite{zhao2021invertible}, image hiding \cite{jing2021hinet}, etc. However, to the best of our knowledge, there has been no work trying to apply INN to the task of optical aberration correction. 

\section{Method}
In this section, we first introduce the simulation method of degraded images based on ray tracing in Section \ref{sec:3.1}. Second, we illustrate the overall architecture of our invertible aberration correction neural network in Section \ref{sec:3.2}. And in Section \ref{sec:3.3}, we elaborate on the composition of the loss function.

\subsection{Raytracing Based Imaging Simulation}
\label{sec:3.1}

\begin{figure}[htbp]
\centering\includegraphics[width=14cm]{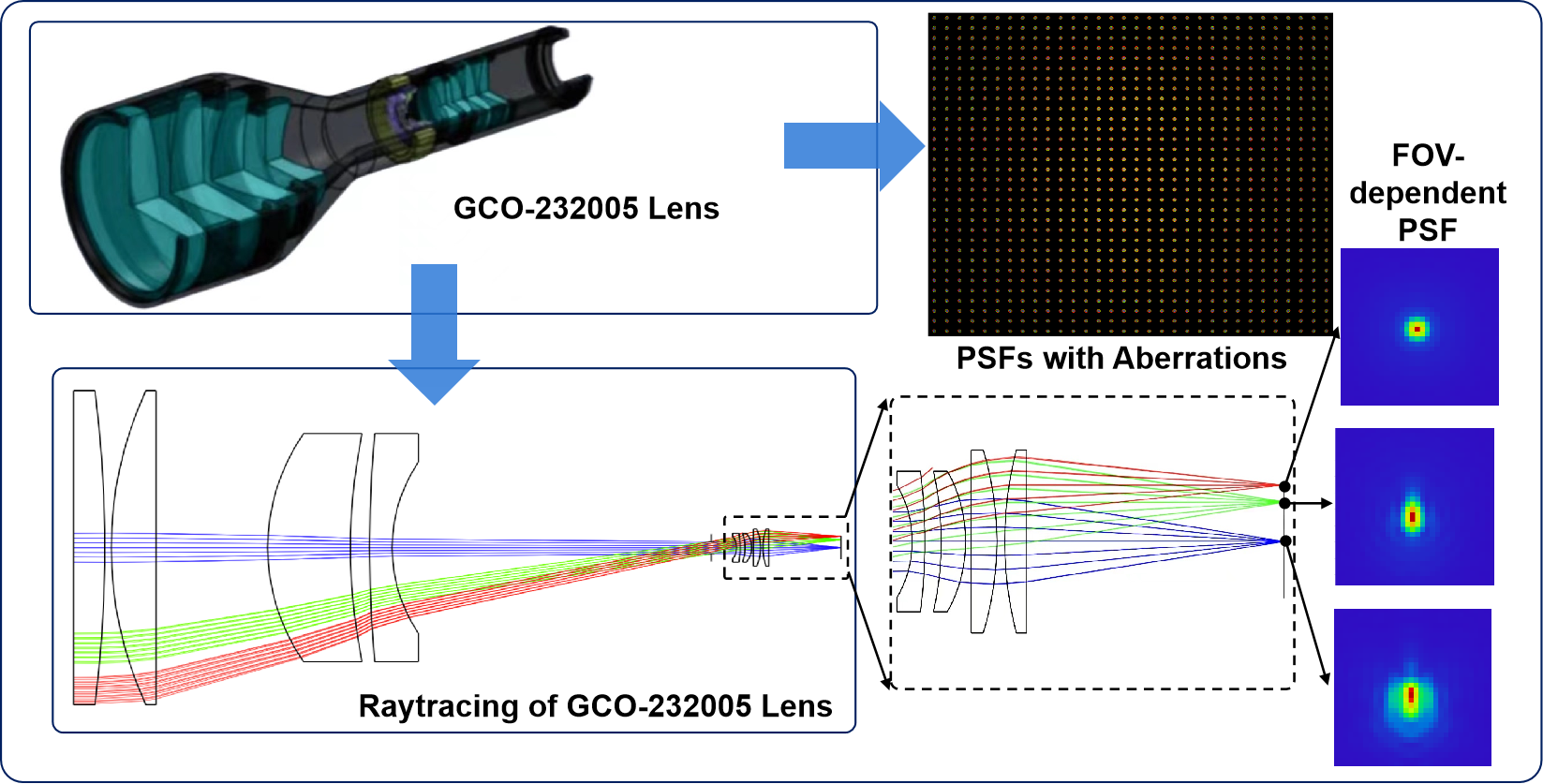}
\caption{The optical system structure and PSFs of GCO-232005 lens. The upper-left shows the cross-sectional view. The upper-right shows the PSFs of the GCO-232005 lens. The lower-left shows the raytracing process in the 2-dimensional plane. The lower-right shows the PSFs for three representative FOVs, illustrating that the PSF varies with FOV.}
\label{Layout-2D}
\end{figure}

The main problem of the supervised aberrations correction algorithm based on deep learning is the lack of real paired datasets. Established methods ignore the underlying optical systems and simply synthesize the degradation caused by optical aberration with a Gaussian degenerate kernel, leading to a large gap between the synthetic paired data and the real data. For optical systems with large fields of view and large apertures, the commonly used Gaussian degenerate kernel causes inaccurate simulation since the actual point spread functions (PSFs) are spatially varying across the field of view (FOV).

The recent work \cite{chen2021optical} has proposed using the imaging simulation method without shooting or registration operation to solve the above problems. This method is easy to migrate to different optical systems. We adopt this approach\cite{chen2021optical} and introduce a distance to the focal plane as one of the simulation inputs to generate synthetic degraded images with varying-degree aberrations.

Our imaging simulation process consists of two steps: first, we calculate patch-wise PSFs with accurate raytracing; second, assuming that the degradation in a local region of a natural image is similar, The sharp image patches are convolved with the patch-wise PSFs to simulate the degradation process. Assuming that the degradation degree of the image patch $I_{degraded}(i,j,d)$ is consistent when shooting at the same object distance, the degradation process can be modeled as follows: 
\begin{equation}\label{eq1}
I_{degraded}(i,j,d) = I_{sharp}(i,j) \otimes k(i,j,d) + n(i,j)
\end{equation}
where $(i,j)$ indicates the spatial coordinates of the patch $I_{degraded}(i,j,d)$. $d$ is the distance to the focal plane. $I_{sharp}(i,j)$ is the latent sharp image of $I_{degraded}(i,j,d)$. $\otimes$ is the operation of convolution. $k(i,j,d)$ is the normalized point spread function, representing the energy diffusion caused by the aberrations of the optical system. $n(i,j)$ models the random noise introduced in the imaging process, and noise can be approximated as the well-established heteroscedastic Gaussian model \cite{foi2009clipped}.

\textbf{Point spread function calculation.} According to the lens parameters of the optical system used for imaging, the wavefront aberrations and the point spread functions of the optical system are calculated by sequential raytracing. Fig. \ref{Layout-2D} shows the process of raytracing and FOV-dependent PSFs. We define a ray $\mathcal{R}$ = $(x, y, z)$ as follows: 
\begin{equation}
\label{equ_raytracing}
\mathcal{R} = \mathcal{O} + t\mathcal{D},t \geq 0.
\end{equation}
where $\mathcal{O}=(x_0, y_0, z_0)$ is the starting point of the ray $\mathcal{R}$, $\mathcal{D}=(d_x, d_y, d_z)$ is the normalized direction vector of the ray $\mathcal{R}$, $\textit{t}$ is the ray marching distance from the starting point $\mathcal{O}$.

The first step of sequential raytracing is to calculate the intersection point of the ray and the surface. For spheric surfaces, the $t$ value for the intersection point is solved analytically. For aspheric surfaces, we define them with sagittal height expression as follows:
\begin{equation}
\label{equ_surface}
z=\frac{c s^2}{1+\sqrt{1-c^2 s^2}}+M_2 s^2+M_4 s^4+\cdots+M_j s^j
\end{equation}
where $z$ is the longitudinal coordinate of a point on the surface, $c$ is the curvature of the spherical part, $s = \sqrt{x^2 + y^2}$ is the distance from the point to $z$ axis. $M_2$, $M_4$ and $M_j$ are the coefficients of higher order terms. The value of $t$ for intersections can be solved by plunging Eq. \ref{equ_raytracing} into Eq. \ref{equ_surface}. Since Eq. \ref{equ_surface} contains high-order terms, the coordinate of the intersection point can only be calculated by multiple iterations numerically\cite{chen2021optical}.

The second step of sequential raytracing is to calculate the refraction direction of the ray. Given the refractive indices $n_1$ and $n_2$ on both sides of the refraction surface and the incident angle $I$ of the incident ray, we use Snell's law to calculate the direction of the refracted ray.


The wavefront aberrations are the deviation of the actual wavefront and the ideal wavefront, which are expressed by the optical path difference. When calculating the optical path difference for a given ray, the ray starts from the object plane, reaches the image plane, and then reversely traces from the image plane back to the reference sphere at the exit pupil. The complex pupil function can be constructed by combining the phase information of the optical path difference and the amplitude information formed by the exit pupil. The pupil function can be expressed as:
\begin{equation}
\mathcal{P}(x, y) = A(x, y) e^{j \frac{2\pi}{\lambda} \phi(x, y)}
\end{equation}
where $(x, y)$ represents the pupil plane coordinates, $A(x, y)$ is the complex amplitude distribution of the exit pupil surface, $\phi(x, y)$ is the optical path difference between the ray at the exit pupil and the chief ray.

The point spread function is the spot formed by the rays from point light sources after passing through the optical system. The amplitude spread function is the Fourier transform of the pupil function $\mathcal{P}(x, y)$. The amplitude spread function can be expressed as:

\begin{equation}
h(u, v) = \int\limits_{-\infty}^{\infty} \int\limits_{-\infty}^{\infty} \mathcal{P}(x, y) e^{-j 2 \pi(u x+v y)} d x d y
\end{equation}
where $(u, v)$ represents the image plane coordinates. The point spread function is the squared magnitude of the amplitude spread function $h(u, v)$.

\textbf{Patch-wise spatial domain convolution.} First, we segment the image into 32 $\times$ 32 uniform patches. These image patches are respectively convolved with PSFs of the corresponding center FOVs in the spatial domain to simulate imaging. See Eq. \ref{eq1} for the specific operations. Then, we splice the degraded image patches together. Finally, we multiply pixel values of the degraded image by the relative illuminance coefficient at the corresponding FOV, which can be obtained according to the pixel position. It should be noted that two additional operations in this process: On the one hand, the PSF needs to be normalized in advance to keep the energy of the image unchanged; On the other hand, to ensure that the smoothness of the image is not affected by patch-wise convolution, the edges of the patches need to be interpolated.

\subsection{Invertible Aberration Correction Architecture}
\label{sec:3.2}
\begin{figure}[htbp]
\centering\includegraphics[width=13cm]{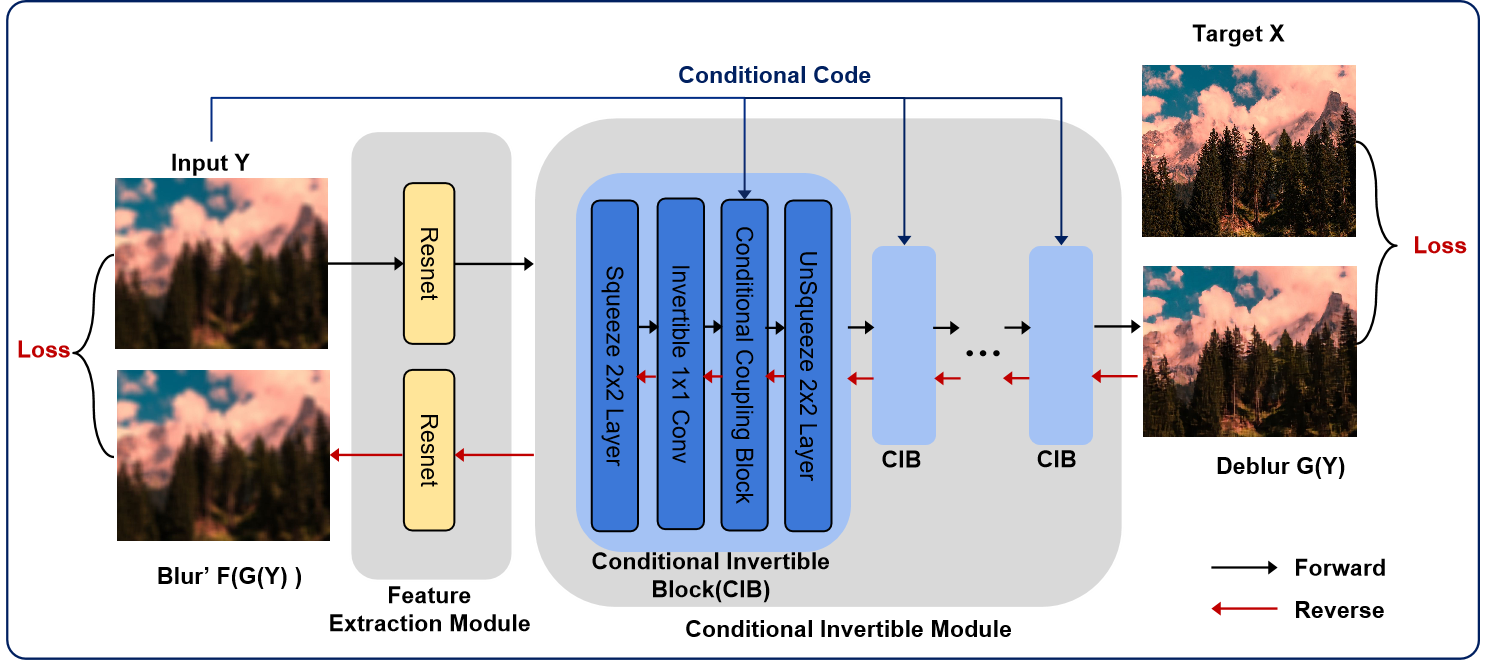}
\caption{\textbf{Overview of the proposed conditional invertible neural network with feature extraction for variable-degree optical aberrations correction.} In the forward process(black arrow), the input degraded image passes through a conditional invertible neural network composed of a feature extraction module and 12 conditional invertible blocks to obtain a sharp image $\mathcal{G}(\mathbf{Y})$. In the reverse process(red arrow), we reverse $\mathcal{G}(\mathbf{Y})$ into the network to obtain the degraded image $\mathcal{F}(\mathcal{G}(\mathbf{Y}))$. Under the joint forward and reverse process, the information of the image is kept as much as possible, making details of the restored image clearer.}
\label{architecture}
\end{figure}

We design a conditional invertible neural network to conduct the aberration correction. Fig. \ref{architecture} shows the overall architecture, which consists of a feature extraction module and a conditional invertible module to correct optical aberrations of variable degrees.

\begin{figure}[htbp]
\centering\includegraphics[width=12cm]{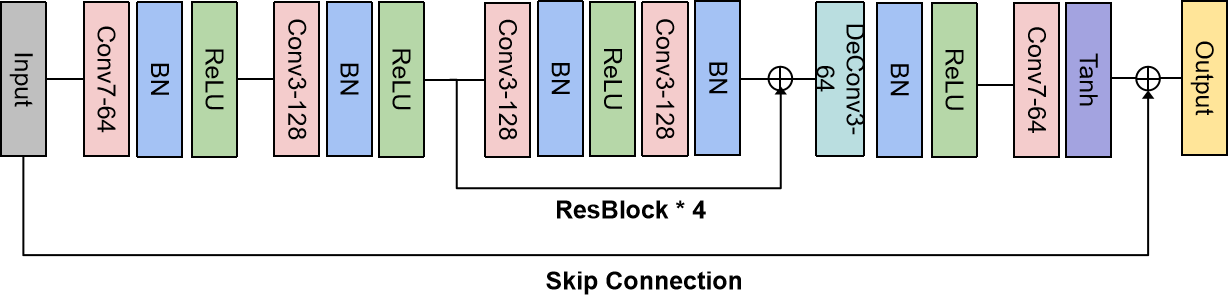}
\caption{\textbf{The detailed architecture of feature extraction module.} “Conv7-64” means that the convolution kernel size of this layer is 7 $\times$ 7, and the number of convolution kernels is 64.}
\label{ResNet}
\end{figure}
\textbf{Feature Extraction Module.} The design of an invertible neural network (INN) needs to ensure strict reversibility, thus INN usually has limited the nonlinear transformation ability. Thus, we add a feature extraction module in front of the INN to improve the nonlinear transformation ability. The feature extraction module is based on multi-scale ResBlocks \cite{he2016deep}, including up-sampling and down-sampling processes. The details are shown in Fig. \ref{ResNet}. It should be noted that the weights of the feature extraction module of the forward and reverse processes are not shared.

\textbf{Conditional Invertible Module.} The conditional invertible module is composed of $k$ conditional invertible blocks, where $k$ is set to 12. Each conditional invertible block consists of the squeeze operation, invertible 1 $\times$ 1 convolution, conditional affine coupling layer, and unsqueeze operation. These operations are invertible, thus the entire INN is completely invertible. Next, we elaborate on the components of the conditional invertible block.

\begin{figure}[htbp]
\centering\includegraphics[width=6cm]{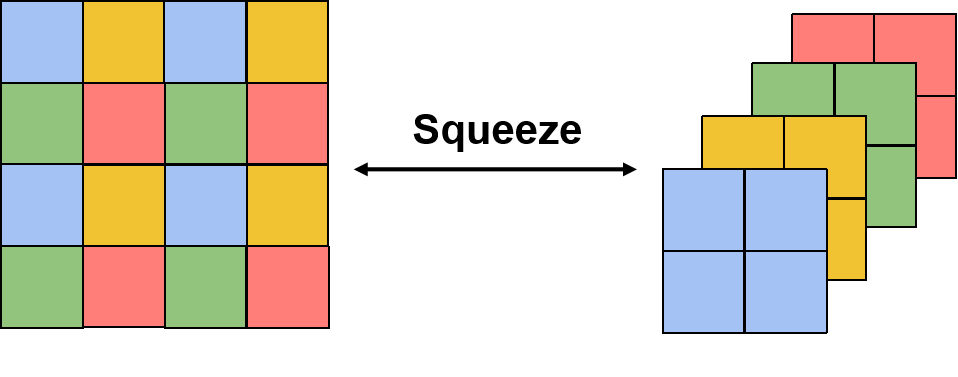}
\caption{The squeeze operation resets a 4 $\times$ 4 $\times$ 1 tensor (on the left) to a 2 $\times$ 2 $\times$ 4 tensor (on the right).}
\label{squeeze}
\end{figure}

\textbf{Squeeze and unsqueeze.} The squeeze operation \cite{dinh2016density} is similar to the convolution operation in CNN, reducing the size of the feature map and increasing the channel of the feature map to capture correlation and structure over a greater spatial distance. Unlike the convolution operation, the squeeze operation extracts features according to the checkerboard pattern to ensure reversibility, as shown in Fig. \ref{squeeze}. The squeeze operation increases the channel dimension while retaining the local correlation of the image. The unsqueeze operation is the inverse process of the squeeze operation to recover the original size of the feature map.

\textbf{Invertible 1 $\times$ 1 convolution.} The invertible 1 $\times$ 1 convolution\cite{kingma2018glow} is a learnable convolution. It fuses information between different feature channels. This allows more interaction and fusion between the information from different incoming data flows.

\begin{figure}[htbp]
	\centering
	\begin{subfigure}{0.47\linewidth}
		\centering
		\includegraphics[width=0.50\linewidth]{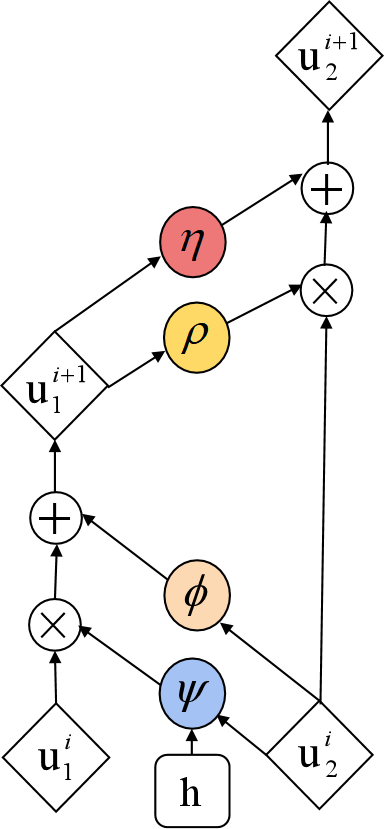}
		\subcaption{Forward Process}
	\end{subfigure}
	\centering
	\begin{subfigure}{0.47\linewidth}
		\centering
		\includegraphics[width=0.49\linewidth]{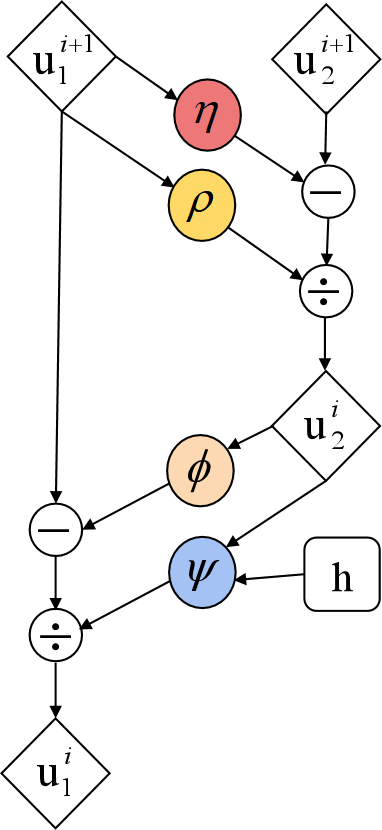}
		\subcaption{Reverse Process}
	\end{subfigure}
	\caption{The conditional affine coupling layer for forward and reverse process}.
    \label{conditional-affine-coupling-layers}
\end{figure}

\textbf{Conditional affine coupling layers.} The schematic diagram of the proposed layers is shown in Fig. \ref{conditional-affine-coupling-layers}. The conditional affine coupling layer\cite{ardizzone2019guided} adds coded condition variables to the affine coupling layer\cite{dinh2016density,dinh2015nice} to improve the efficiency of the flow model. Here, the conditional code $h$ represents the aberrations degree of the image, specifically referring to the object distance to the focal plane. We use binary mode to encode the distance. The images in our dataset have 101 different distances, so the length of the binary code is 7-bit. For the i-th affine coupling layer, the input $u^i$ is divided into $u^i_1$ and $u^i_2$ along the channel direction, and then they undergo augmented affine transformation\cite{dinh2016density,xiao2020invertible}:
\begin{equation}\label{forward}
\begin{aligned}
&u_1^{i},u_2^{i}=Split(u^{i})\\
&u_1^{i+1}=u_1^i \odot \exp \left(\psi\left(u_2^i,h\right)\right)+\phi\left(u_2^i\right) \\
&u_2^{i+1}=u_2^i \odot \exp \left(\rho\left(u_1^{i+1}\right)\right)+\eta\left(u_1^{i+1}\right)\\
&u^{i+1}=Concat(u_1^{i+1},u_2^{i+1})
\end{aligned}
\end{equation}

The Eq. \ref{forward} corresponds to a forward process, and the outputs [$u^{i+1}_1$, $u^{i+1}_2$]  are concatenated again and passed to the next affine coupling block. In the reverse process, only addition (+) and multiplication ($\times$) operations are reversed to subtraction (-) and division (/), while the internal transformation functions($\psi()$, $\phi()$, $\rho()$, $\eta()$) do not need to be reversible, and they can be represented by arbitrary neural networks. we employ a multi-scale residual concatenated convolutional block, which is a simplified version of the feature enhancement module. Specifically, we changed the 64 convolution kernels in the first pink block "Conv7-64" to 32 convolution kernels("Conv7-32") in Fig. \ref{ResNet}, and we reduced the middle "ResBlock $\times$ 4" to "ResBlock $\times$ 2". When the output $u^{i+1}$ is given, the corresponding reverse process can be expressed as:
\begin{equation}
\begin{aligned}
&u_1^{i+1},u_2^{i+1}=Split(u^{i+1})\\
&u_2^i=\left(u_2^{i+1}-\eta\left(u_1^{i+1}\right)\right) \odot \exp \left(-\rho\left(u_1^{i+1}\right)\right) \\
&u_1^i=\left(u_1^{i+1}-\phi\left(u_2^i\right)\right) \odot \exp \left(-\psi\left(u_2^i,h\right)\right)\\
&u^{i}=Concat(u_1^{i},u_2^{i})
\end{aligned}
\end{equation}

\subsection{Loss Function}
\label{sec:3.3}
We optimize the proposed invertible aberration correction neural network end-to-end with the following loss function:
\begin{equation}
\begin{aligned}
\mathcal{L}_{total} =  \lambda_{1}\mathcal{L}_{forward}(\mathcal{G}(\mathbf{Y}),\mathbf{X}) + \lambda_{2}\mathcal{L}_{reverse}(\mathcal{F}(\mathcal{G}(\mathbf{Y})),\mathbf{Y}) \\+ \lambda_{3}\mathcal{L}_{edge}(\mathcal{G}(\mathbf{Y}),\mathbf{X}) + \lambda_{4}\mathcal{L}_{perceptual}(\mathcal{G}(\mathbf{Y}),\mathbf{X})
\end{aligned}
\end{equation}
where $\lambda_{1}$, $\lambda_{2}$, $\lambda_{3}$, $\lambda_{4}$ are hyperparameters, which control the importance of different loss items respectively. They are empirically set as $\lambda_{1}$ = 1, $\lambda_{2}$ = 0.5,  $\lambda_{3}$ = 0.05 and $\lambda_{4}$ = 0.02. $\mathcal{G}$ stands for the forward transformation from degraded images to sharp images. $\mathcal{F}$ is the inverse process of $\mathcal{G}$. $\mathbf{X}$ is the reference sharp image. $\mathbf{Y}$ is the input image with optical aberrations. 

\textbf{Forward Loss.} This loss is applied in the forward process to eliminate the optical aberrations of the image, such that the output content is close to the content of the reference. The $l_1$ norm regularized pixel level loss function is used as it provides better quality than other norms, like the $l_2$ norm.
\begin{equation}
\mathcal{L}_{forward}(\mathcal{G}(\mathbf{Y}),\mathbf{X})=\|\mathcal{G}(\mathbf{Y})-\mathbf{X}\|_1
\end{equation}

\textbf{Reverse} Loss. The reverse loss makes the learning process more stable and increases the robustness of the neural network. $\mathcal{G}(\mathbf{Y})$ is the output of the degraded image $\mathbf{Y}$ after passing through the forward network, and $\mathcal{F}(\mathcal{G}(\mathbf{Y}))$ is the degraded image after the reverse process. The reverse loss make the content of $\mathcal{F}(\mathcal{G}(\mathbf{Y}))$ close to the initial degraded image $\mathbf{Y}$.
\begin{equation}
\mathcal{L}_{reverse}(\mathcal{F}(\mathcal{G}(\mathbf{Y}),\mathbf{Y})=\|\mathcal{F}(\mathcal{G}(\mathbf{Y}))-\mathbf{Y}\|_1
\end{equation}

\textbf{Edge Loss.} The edge loss can take high frequency texture structure information into account, and improve the details of hyper-segmented images.
\begin{equation}
\mathcal{L}_{edge}(\mathcal{G}(\mathbf{Y}),\mathbf{\mathbf{X}}) = \left\|\Delta\left(\mathcal{G}(\mathbf{Y})\right)-\Delta(\mathbf{X})\right\|_1
\end{equation}
where $\Delta$ denotes the Laplacian operator.

\textbf{Perceptual Loss.} Perceptual Loss\cite{johnson2016perceptual} measures the difference between two images by features extracted from the benchmark VGG model \cite{simonyan2014very}. It enhances the perceptual similarity between the generated image and the reference image, thus helping to produce a more realistic image.
\begin{equation}
\mathcal{L}_{perceptual}(\mathcal{G}(\mathbf{Y}),\mathbf{X})=\frac{1}{C_m H_m W_m}\|\varphi_m(\mathbf{X})-\varphi_m(\mathcal{G}(\mathbf{Y}))\|_1
\end{equation}
where $m$ represents the m-th layer; $C_m$, $H_m$ and $W_m$ stand for the number of channels, height, and width of the feature maps, respectively; $\varphi_m(\mathbf{X})$ represents the feature response of the sharp $\mathbf{X}$ at the m-th layer, and $\varphi_m(\mathcal{G}(\mathbf{Y}))$ represents the feature response of $\mathcal{G}(\mathbf{Y})$ at the m-th layer. In this work, we use the tenth ($m=10$) convolutional layer of the pretrained VGG-19 network to extract features.

\section{Experiments}

This section first introduces the details of the experimental setup and then qualitatively and quantitatively compares our proposed method with state-of-the-art methods on both the synthetic data and the real data. Finally, we further conducted ablation studies and provide an analysis of the results.

\subsection{Experimental Settings}
\label{sec:4.1}
We use the DIV2K \cite{agustsson2017ntire} dataset, consisting of 1000 high-quality 2K resolution images. We randomly select a part of the DIV2K dataset and ISO 12233 chart, and we conduct batch imaging simulation to generate paired synthetic datasets. The target optical system is the GCO 232005 optical lens. In imaging simulation, we set the object distance to the focal plane from $-125$ mm to $125$ mm to construct the synthetic dataset, with an interval of $2.5$ mm, corresponding to 101 different degrees of aberrations. For this optical system, when the object distance to the focal plane is more than 80 mm or less than $-80$ mm, the simulated image can be considered as heavily degraded, which brings great challenges to the aberration correction task.

The synthetic dataset contains 6000 image pairs, which are divided into the training set, validation set and test set in a 4:1:1 ratio. We implement the proposed method using PyTorch and train neural networks on two RTX 5000 GPUs with Adam \cite{kingma2015adam} optimizer($\beta_1$ = 0.9, $\beta_2$ = 0.999) for total 150 epochs. The initial learning rate is fixed at 1 × $10^{-4}$, which decays by half every 50 epochs. The training patch size is 256 × 256 and the batch size is 8. We employ random cropping, flipping and rotation to augment the training data as in \cite{park2020multi,mou2022deep}. Generally, it takes around 1.5 days to train a model for 150 epochs. 

We also set up an experimental optical system to capture real aberration data. The optical system consists of GCO 232005 optical lens and an MER-131-210U3C-L CMOS sensor. Same as in our simulation, we capture real images by setting the imaging object distance to the focal plane as values ranging from $-125$ mm to 125 mm with the interval $5.0$ mm. 

\begin{table}
\arrayrulecolor{black}
\begin{center}
\caption{Comparisons of the evaluation metrics on the synthetic dataset and the number of parameters. Note that all methods are trained on the synthetic training set. The best score in each column is highlighted in bold.}
\begin{tabular}{l|lll|l}
\hline Method & PSNR $\uparrow$ & SSIM $\uparrow$ & LPIPS $\downarrow$ & Params$\#$ \\
\hline Input & $21.0932$ & $0.7246$ & $0.4427$ &  \\
\hline 
DeblurGANv2\cite{kupyn2019deblurgan} & $17.5685$ & $0.5712$ & $0.4015$ & $3.31M$ \\
FOV-KPN\cite{chen2021extreme} & $20.6122$ &  $0.7230$ & $0.4047$ & $4.01M$ \\
MIMO-UNet\cite{cho2021rethinking} & $24.9595$ & $\mathbf{0.8493}$ & $\mathbf{0.1956}$ & $6.81M$ \\
MPRNet\cite{zamir2021multi} & $22.3145$ & $0.7639$ & $0.3393$ & $20.13M$ \\
Stripformer\cite{tsai2022stripformer} & $23.6378$ & $0.8104$ & $0.2324$ & $19.71M$ \\
Ours & $\mathbf{25.3004}$ & $0.8432$ & $0.2069$ & $10.16M$ \\
\hline
\end{tabular}
\label{synthetic-table}
\end{center} 
\end{table}

\begin{figure}[htbp]
	\centering
	\begin{subfigure}{0.24\linewidth}
	\setlength{\abovecaptionskip}{0.cm}
	\setlength{\belowcaptionskip}{0.2cm}
		\centering
		\includegraphics[width=\linewidth]{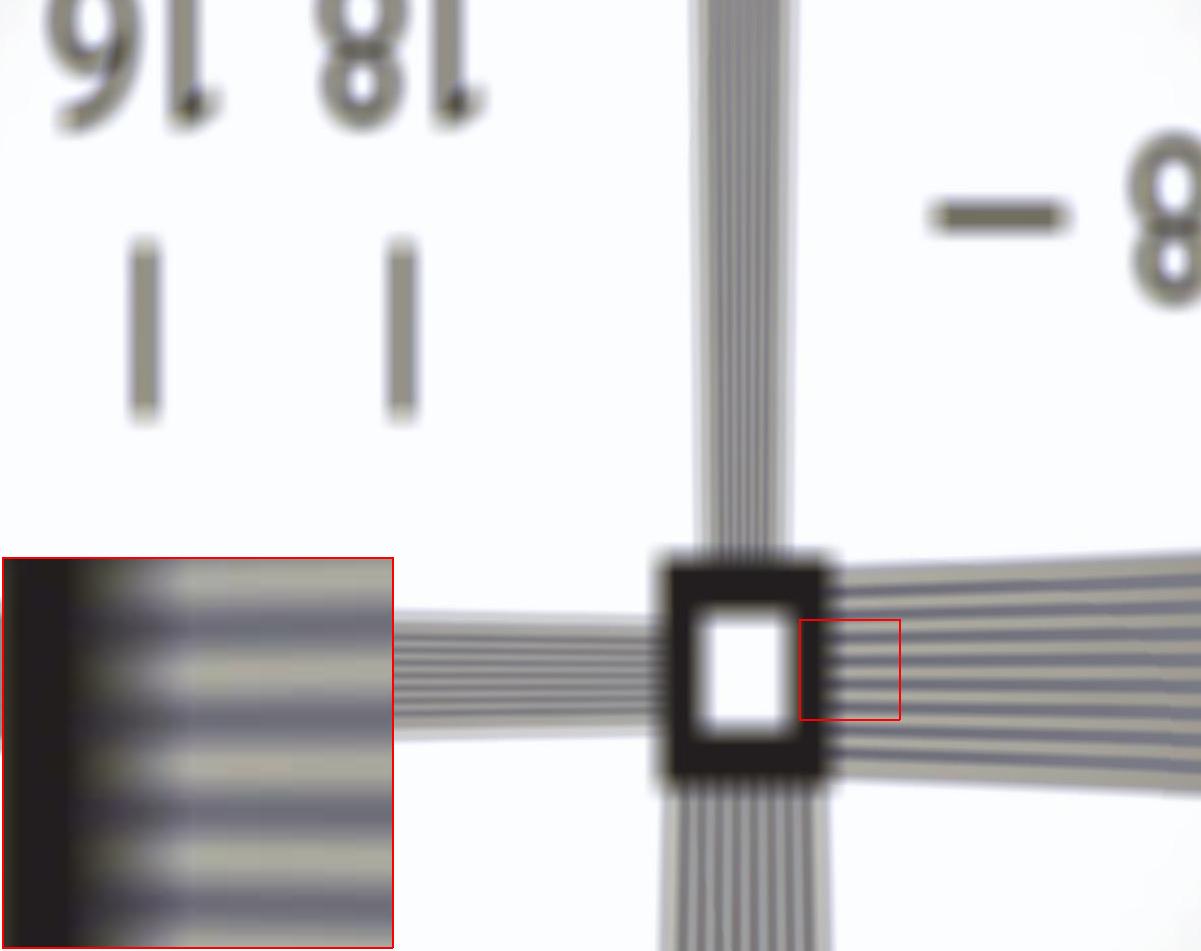}
		\subcaption*{Input(Distance:122.5mm)}
	\end{subfigure}
	\centering
	\begin{subfigure}{0.24\linewidth}
	\setlength{\abovecaptionskip}{0.cm}
	\setlength{\belowcaptionskip}{0.2cm}
		\centering
		\includegraphics[width=\linewidth]{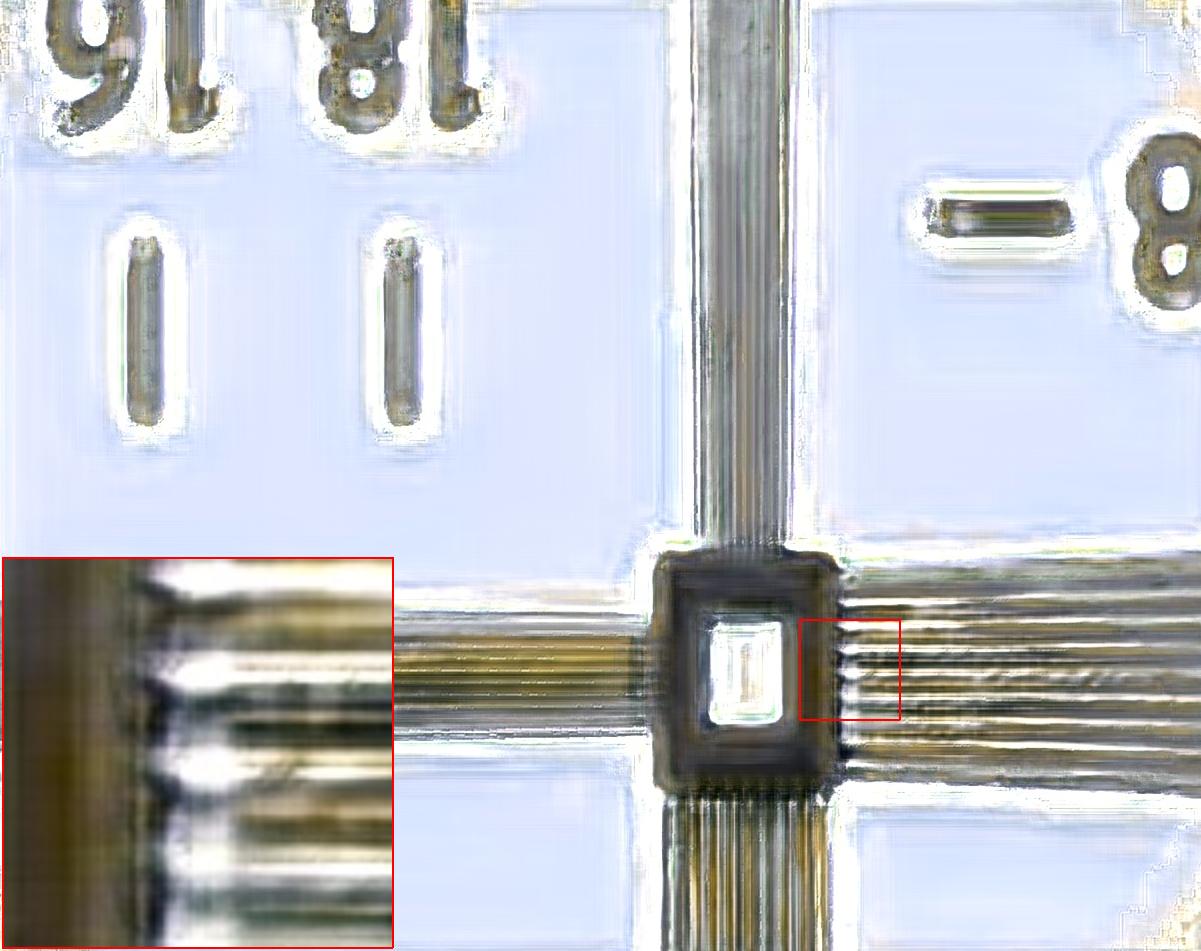}
		\subcaption*{DeblurGANv2}
	\end{subfigure}
	\centering
	\begin{subfigure}{0.24\linewidth}
	\setlength{\abovecaptionskip}{0.cm}
	\setlength{\belowcaptionskip}{0.2cm}
		\centering
		\includegraphics[width=\linewidth]{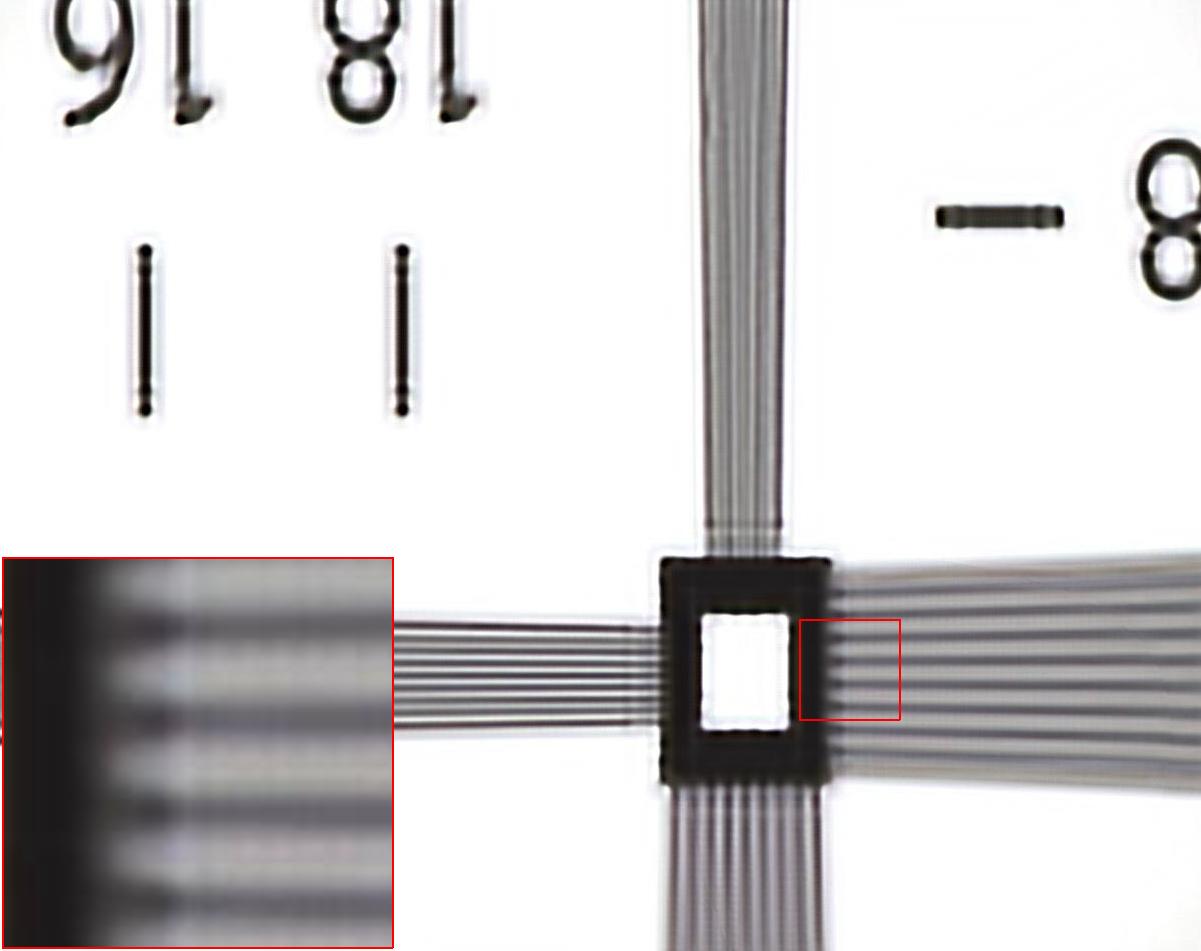}
		\subcaption*{FOV-KPN}
	\end{subfigure}
	\centering
	\begin{subfigure}{0.24\linewidth}
	\setlength{\abovecaptionskip}{0.cm}
	\setlength{\belowcaptionskip}{0.2cm}
		\centering
		\includegraphics[width=\linewidth]{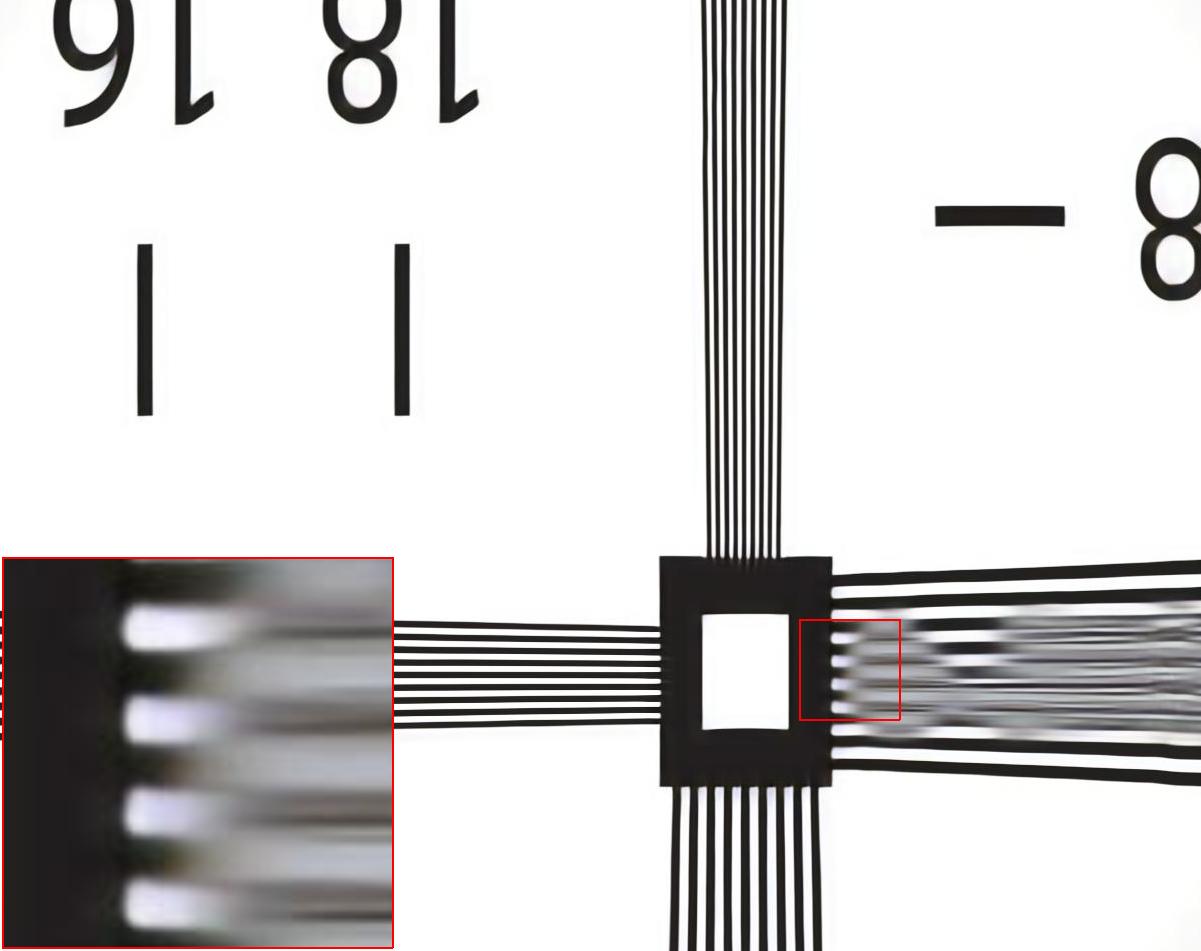}
		\subcaption*{MIMO-UNet}
	\end{subfigure}
	\centering
	\begin{subfigure}{0.24\linewidth}
	\setlength{\abovecaptionskip}{0.cm}
	\setlength{\belowcaptionskip}{0.2cm}
		\centering
		\includegraphics[width=\linewidth]{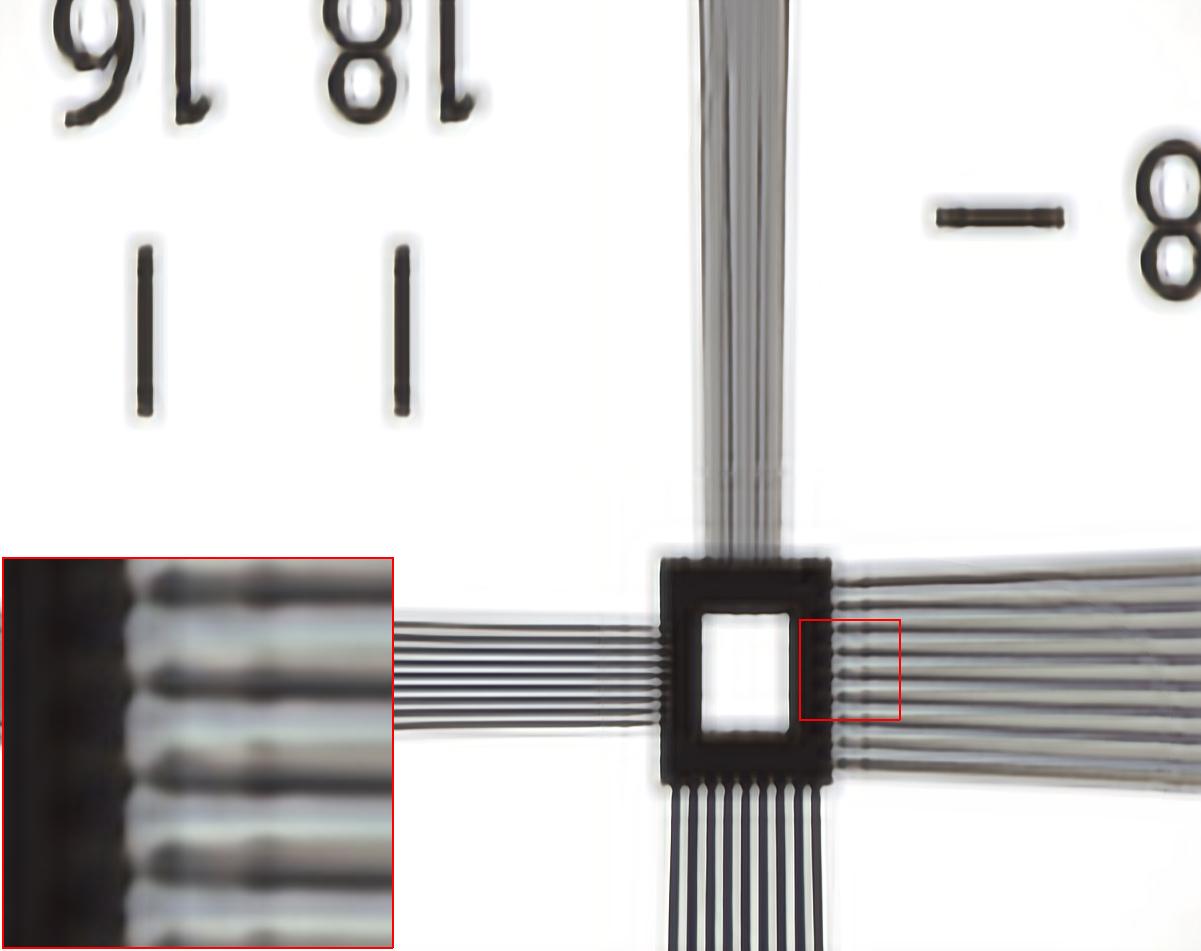}
		\subcaption*{MPRNet}
	\end{subfigure}
	\centering
	\begin{subfigure}{0.24\linewidth}
	\setlength{\abovecaptionskip}{0.cm}
	\setlength{\belowcaptionskip}{0.2cm}
		\centering
		\includegraphics[width=\linewidth]{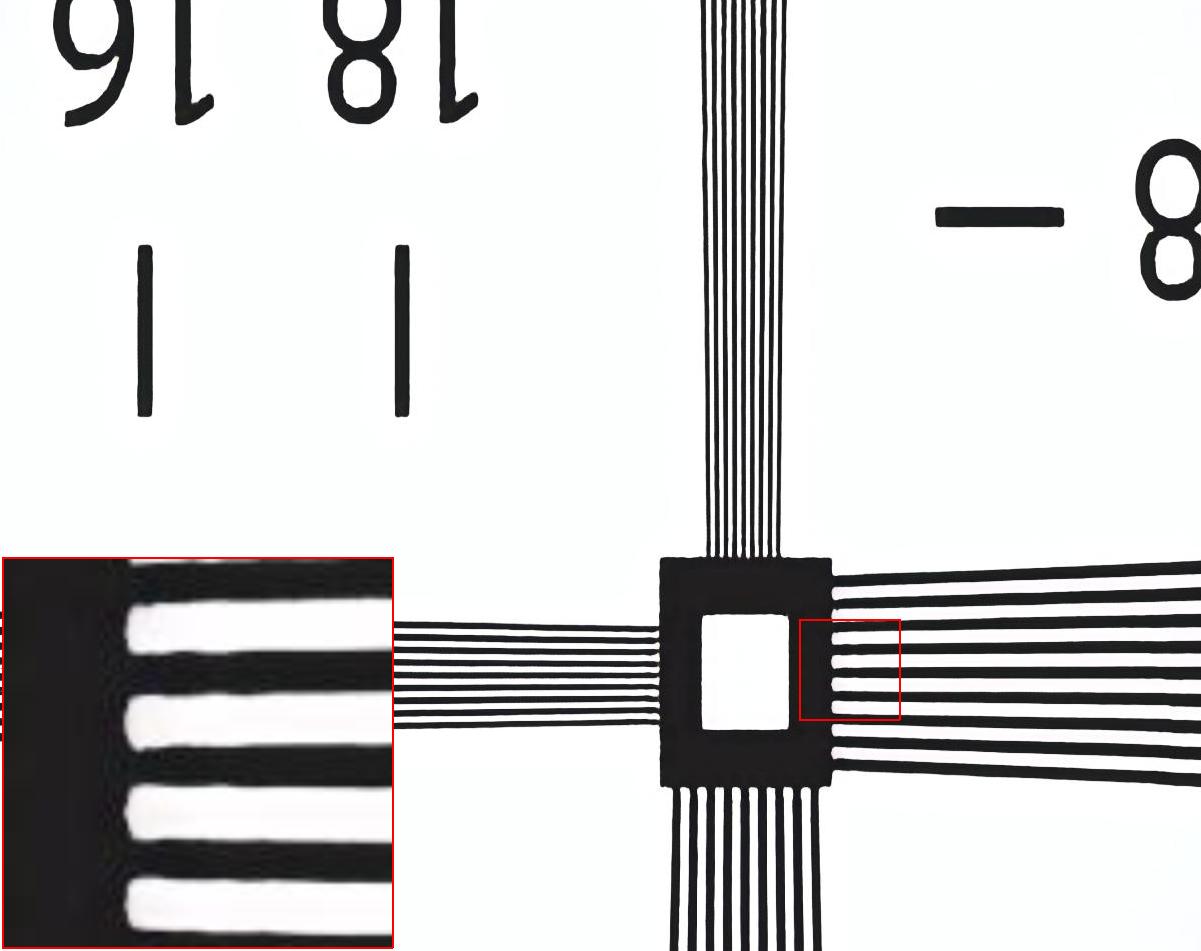}
		\subcaption*{Stripformer}
	\end{subfigure}
	\centering
	\begin{subfigure}{0.24\linewidth}
	\setlength{\abovecaptionskip}{0.cm}
	\setlength{\belowcaptionskip}{0.2cm}
		\centering
		\includegraphics[width=\linewidth]{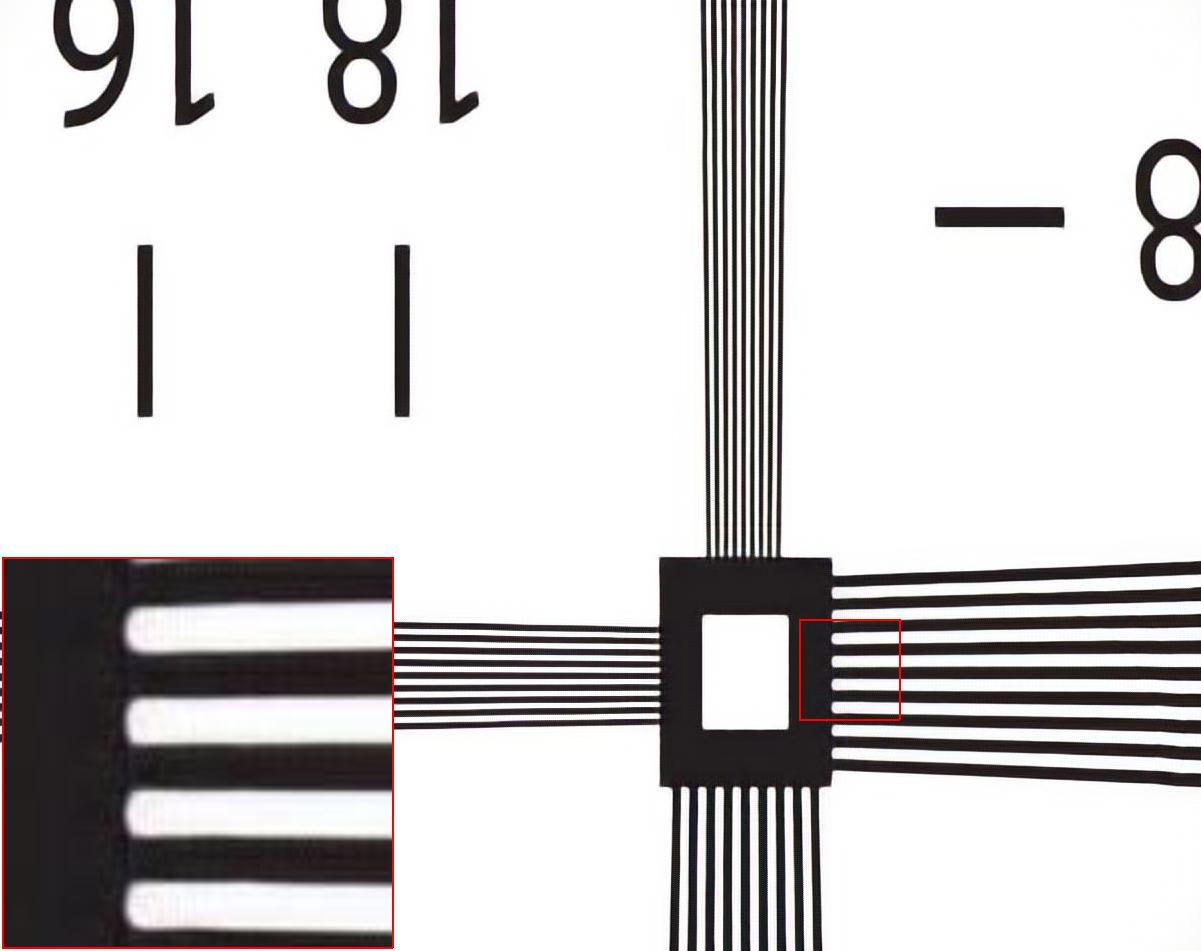}
		\subcaption*{Ours}
	\end{subfigure}
	\centering
	\begin{subfigure}{0.24\linewidth}
	\setlength{\abovecaptionskip}{0.cm}
	\setlength{\belowcaptionskip}{0.2cm}
		\centering
		\includegraphics[width=\linewidth]{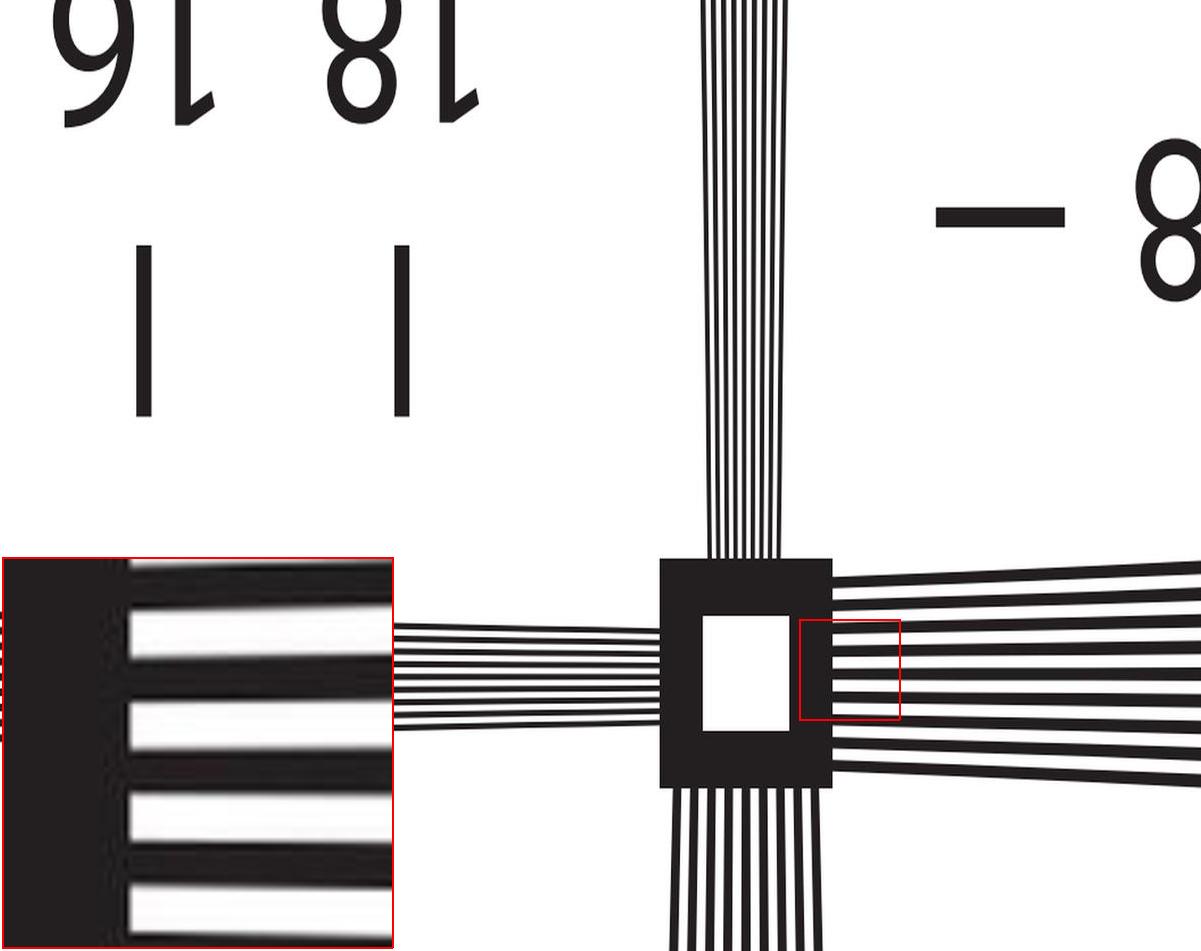}
		\subcaption*{Reference}
	\end{subfigure}
	\centering
	\begin{subfigure}{0.24\linewidth}
	\setlength{\abovecaptionskip}{0.cm}
	\setlength{\belowcaptionskip}{0.2cm}
		\centering
		\includegraphics[width=\linewidth]{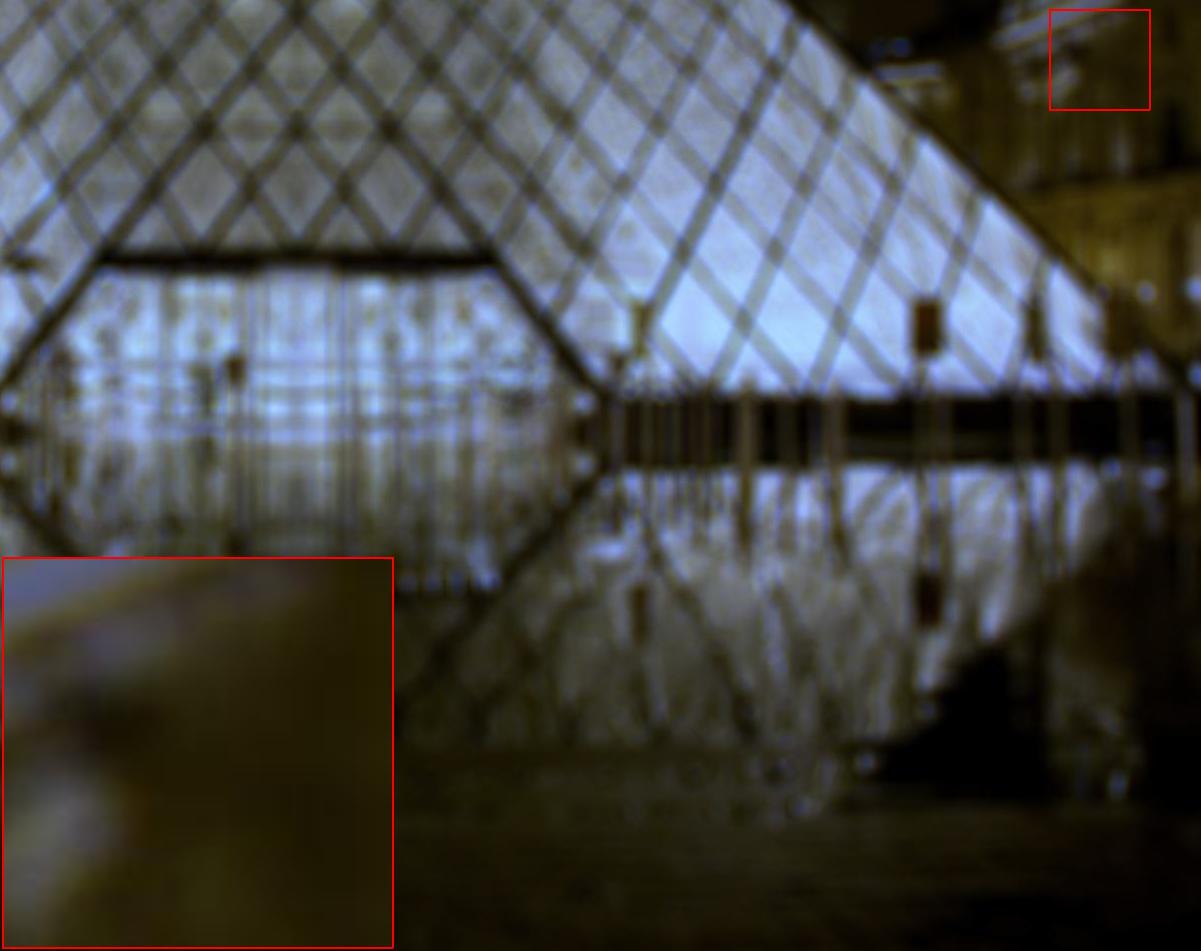}
		\subcaption*{Input(Distance:77.5mm)}
	\end{subfigure}
	\centering
	\begin{subfigure}{0.24\linewidth}
	\setlength{\abovecaptionskip}{0.cm}
	\setlength{\belowcaptionskip}{0.2cm}
		\centering
		\includegraphics[width=\linewidth]{picture/exp-pic2/t0234df82-input.jpg}
		\subcaption*{DeblurGANv2}
	\end{subfigure}
	\centering
	\begin{subfigure}{0.24\linewidth}
	\setlength{\abovecaptionskip}{0.cm}
	\setlength{\belowcaptionskip}{0.2cm}
		\centering
		\includegraphics[width=\linewidth]{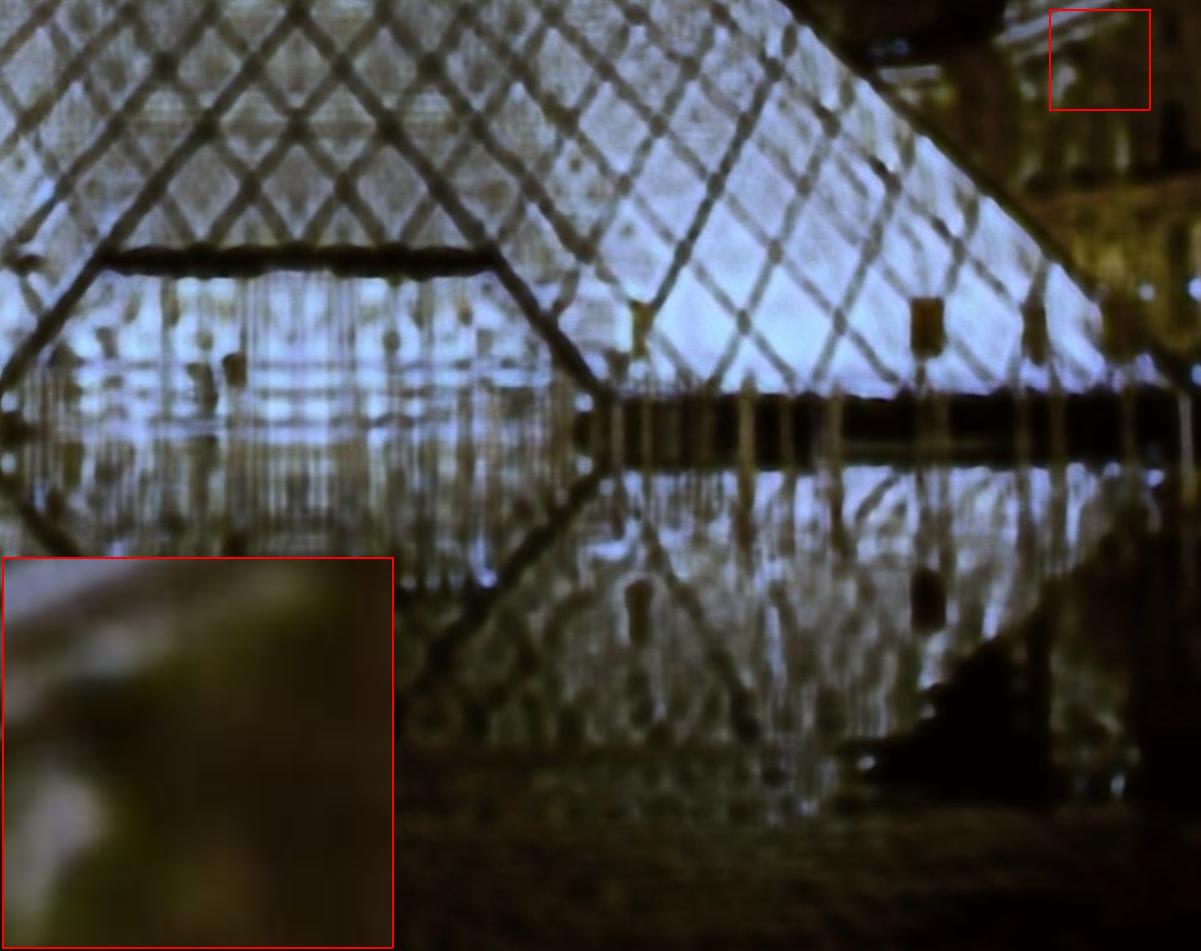}
		\subcaption*{FOV-KPN}
	\end{subfigure}
	\centering
	\begin{subfigure}{0.24\linewidth}
	\setlength{\abovecaptionskip}{0.cm}
	\setlength{\belowcaptionskip}{0.2cm}
		\centering
		\includegraphics[width=\linewidth]{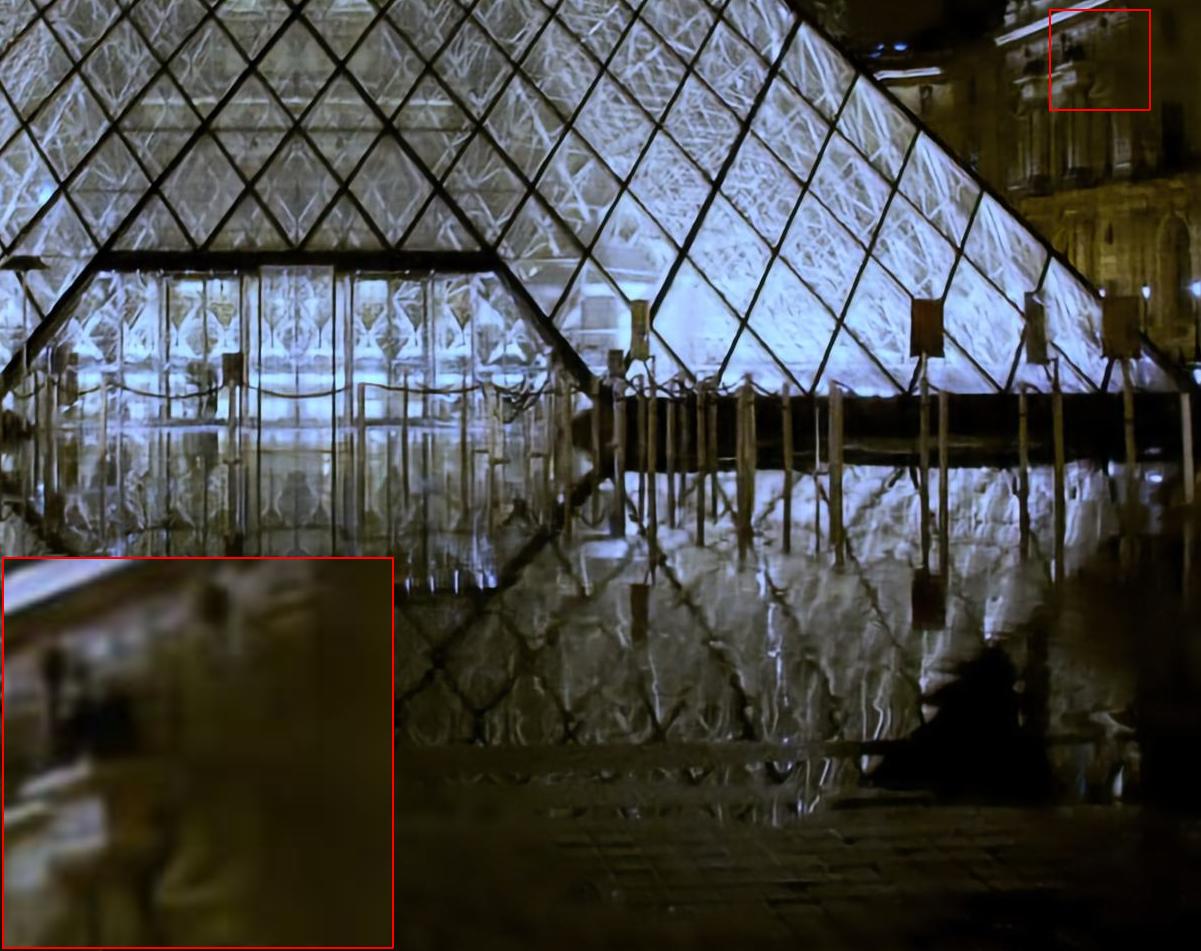}
		\subcaption*{MIMO-UNet}
	\end{subfigure}
	\centering
	\begin{subfigure}{0.24\linewidth}
	\setlength{\abovecaptionskip}{0.cm}
	\setlength{\belowcaptionskip}{0.2cm}
		\centering
		\includegraphics[width=\linewidth]{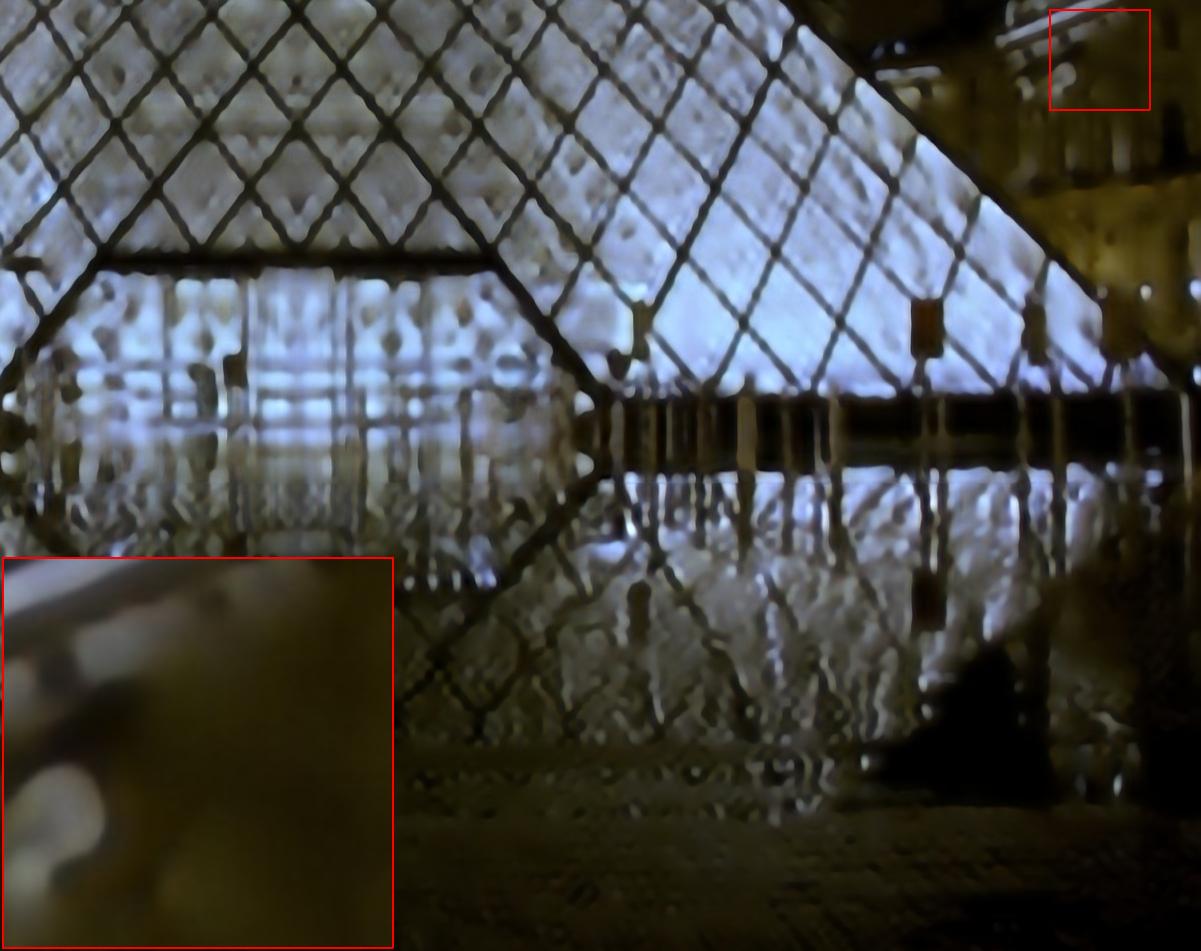}
		\subcaption*{MPRNet}
	\end{subfigure}
	\centering
	\begin{subfigure}{0.24\linewidth}
	\setlength{\abovecaptionskip}{0.cm}
	\setlength{\belowcaptionskip}{0.2cm}
		\centering
		\includegraphics[width=\linewidth]{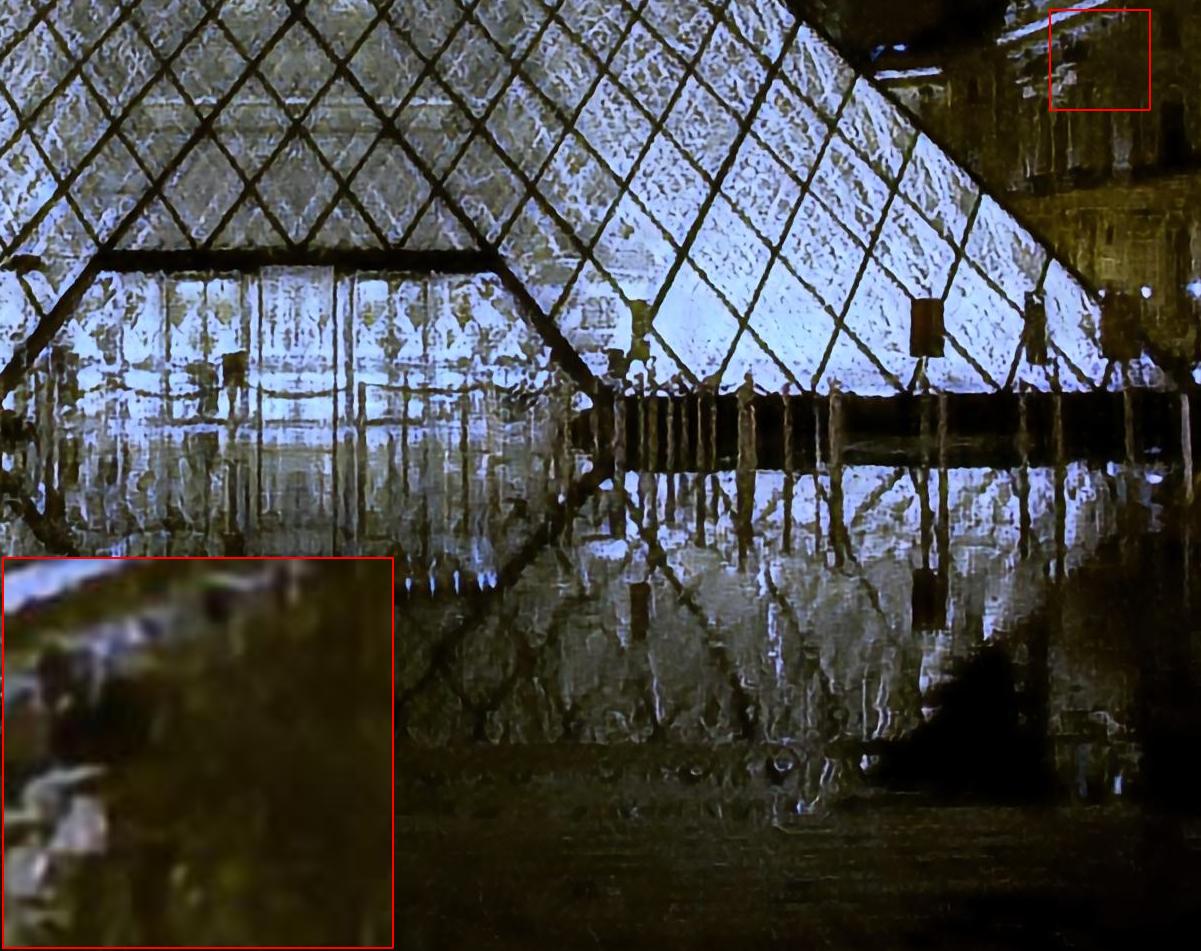}
		\subcaption*{Stripformer}
	\end{subfigure}
	\centering
	\begin{subfigure}{0.24\linewidth}
	\setlength{\abovecaptionskip}{0.cm}
	\setlength{\belowcaptionskip}{0.2cm}
		\centering
		\includegraphics[width=\linewidth]{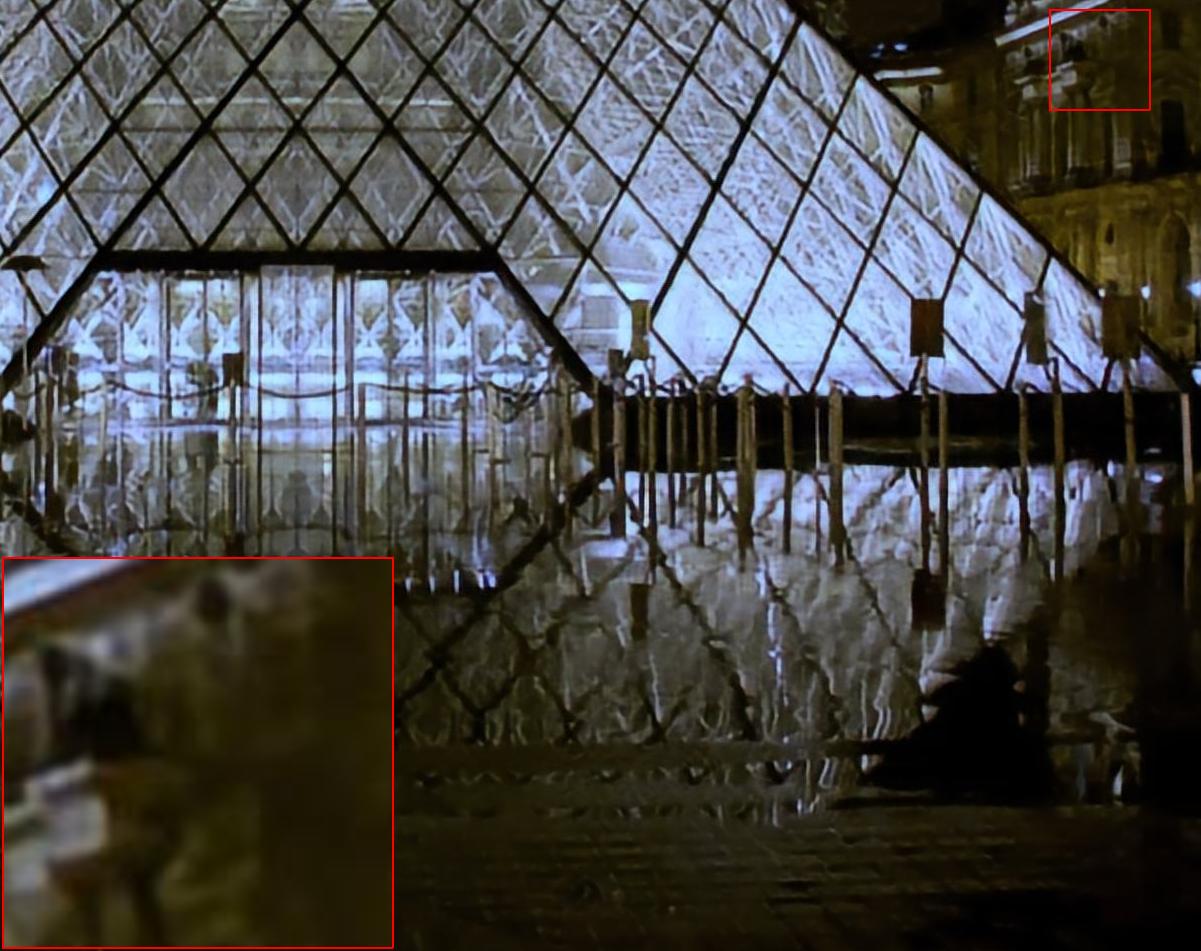}
		\subcaption*{Ours}
	\end{subfigure}
	\centering
	\begin{subfigure}{0.24\linewidth}
	\setlength{\abovecaptionskip}{0.cm}
	\setlength{\belowcaptionskip}{0.2cm}
		\centering
		\includegraphics[width=\linewidth]{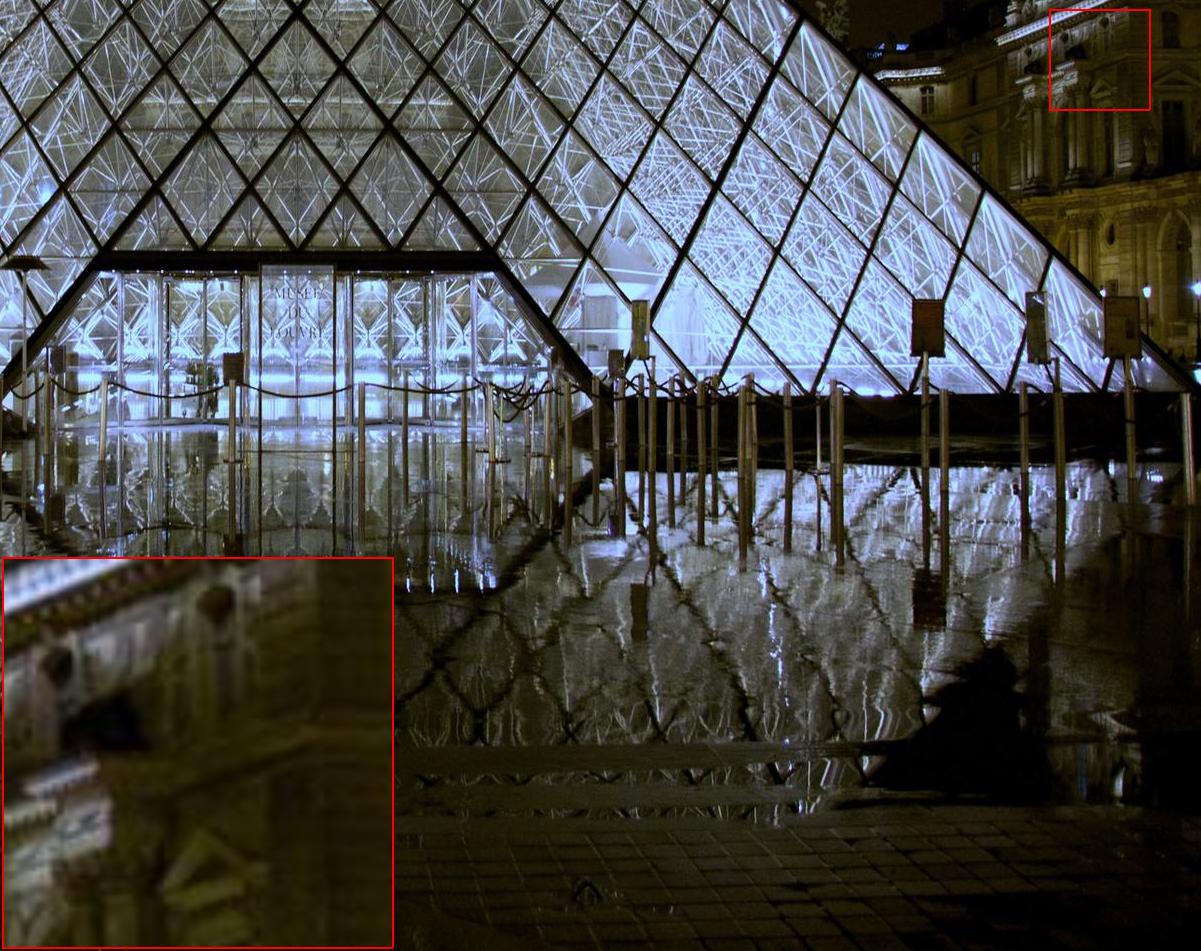}
		\subcaption*{Reference}
	\end{subfigure}
	\centering
	\begin{subfigure}{0.24\linewidth}
	\setlength{\abovecaptionskip}{0.cm}
	\setlength{\belowcaptionskip}{0.2cm}
		\centering
		\includegraphics[width=\linewidth]{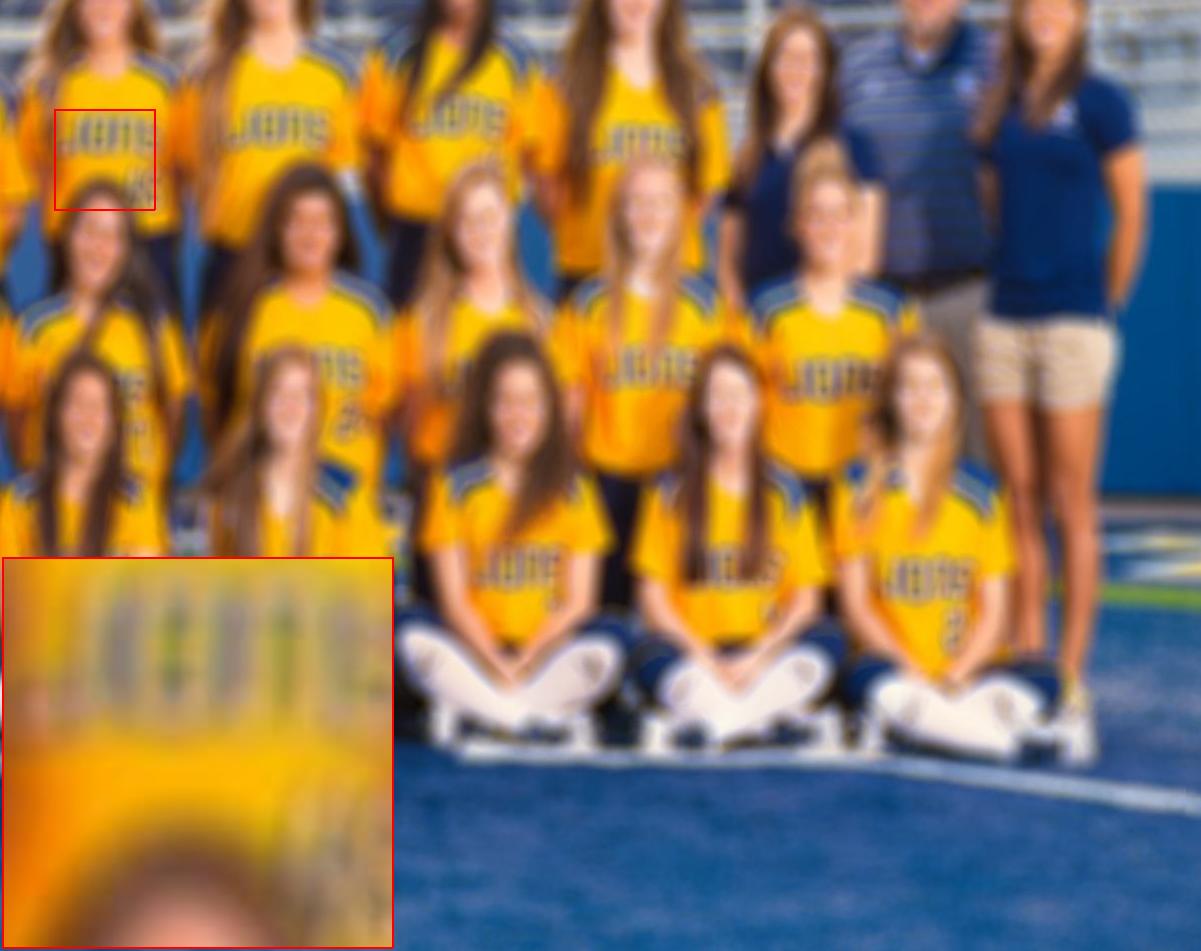}
		\subcaption*{Input(Distance:80.0mm)}
	\end{subfigure}
	\centering
	\begin{subfigure}{0.24\linewidth}
	\setlength{\abovecaptionskip}{0.cm}
	\setlength{\belowcaptionskip}{0.2cm}
		\centering
		\includegraphics[width=\linewidth]{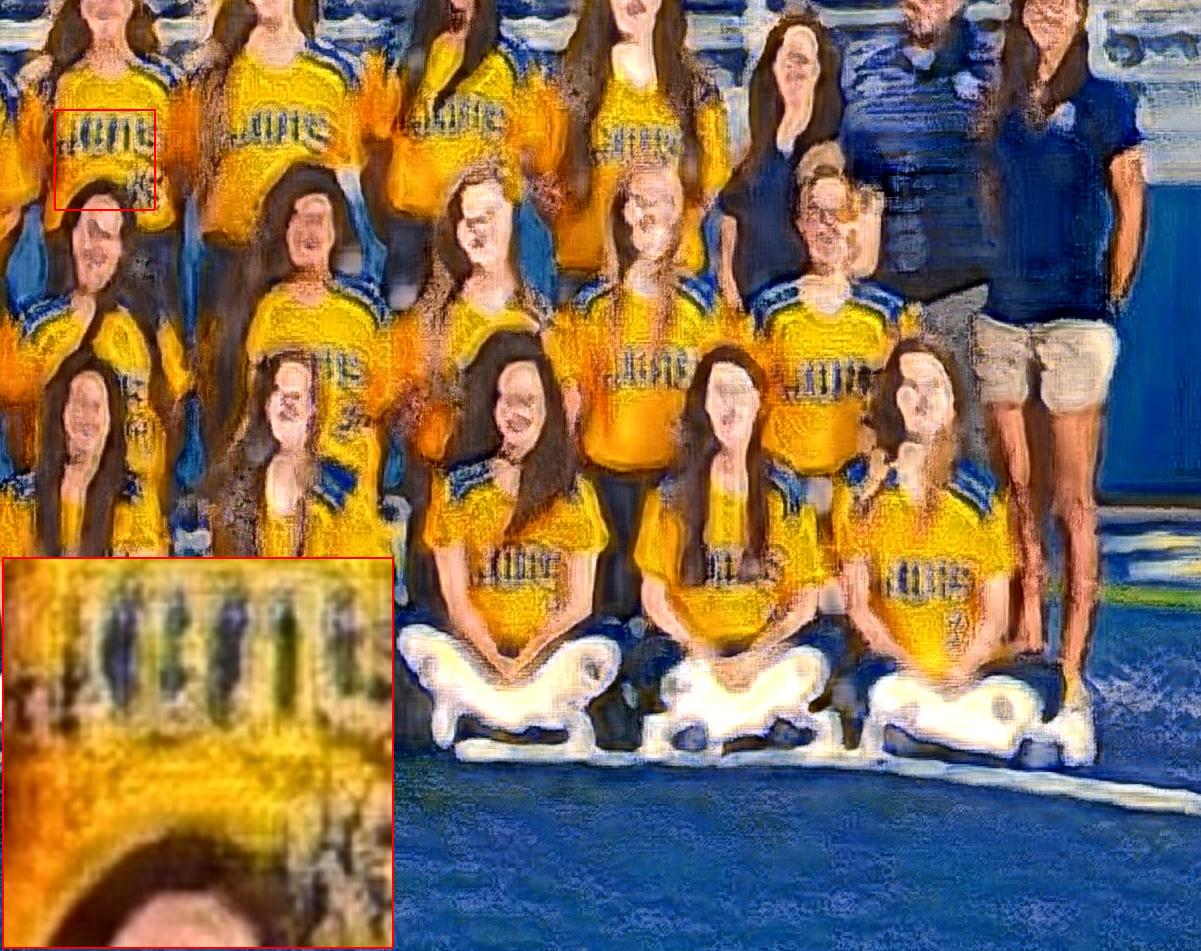}
		\subcaption*{DeblurGANv2}
	\end{subfigure}
	\centering
	\begin{subfigure}{0.24\linewidth}
	\setlength{\abovecaptionskip}{0.cm}
	\setlength{\belowcaptionskip}{0.2cm}
		\centering
		\includegraphics[width=\linewidth]{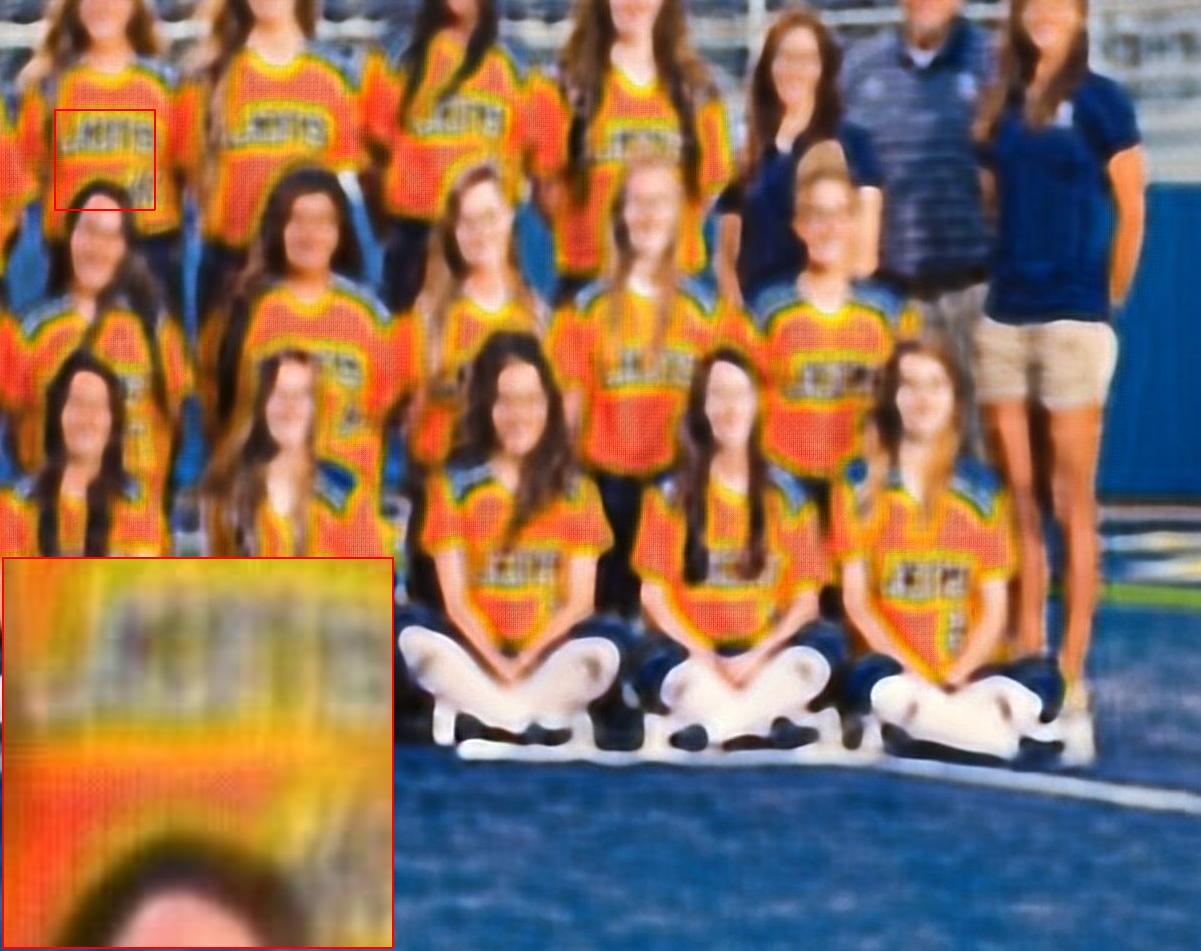}
		\subcaption*{FOV-KPN}
	\end{subfigure}
	\centering
	\begin{subfigure}{0.24\linewidth}
	\setlength{\abovecaptionskip}{0.cm}
	\setlength{\belowcaptionskip}{0.2cm}
		\centering
		\includegraphics[width=\linewidth]{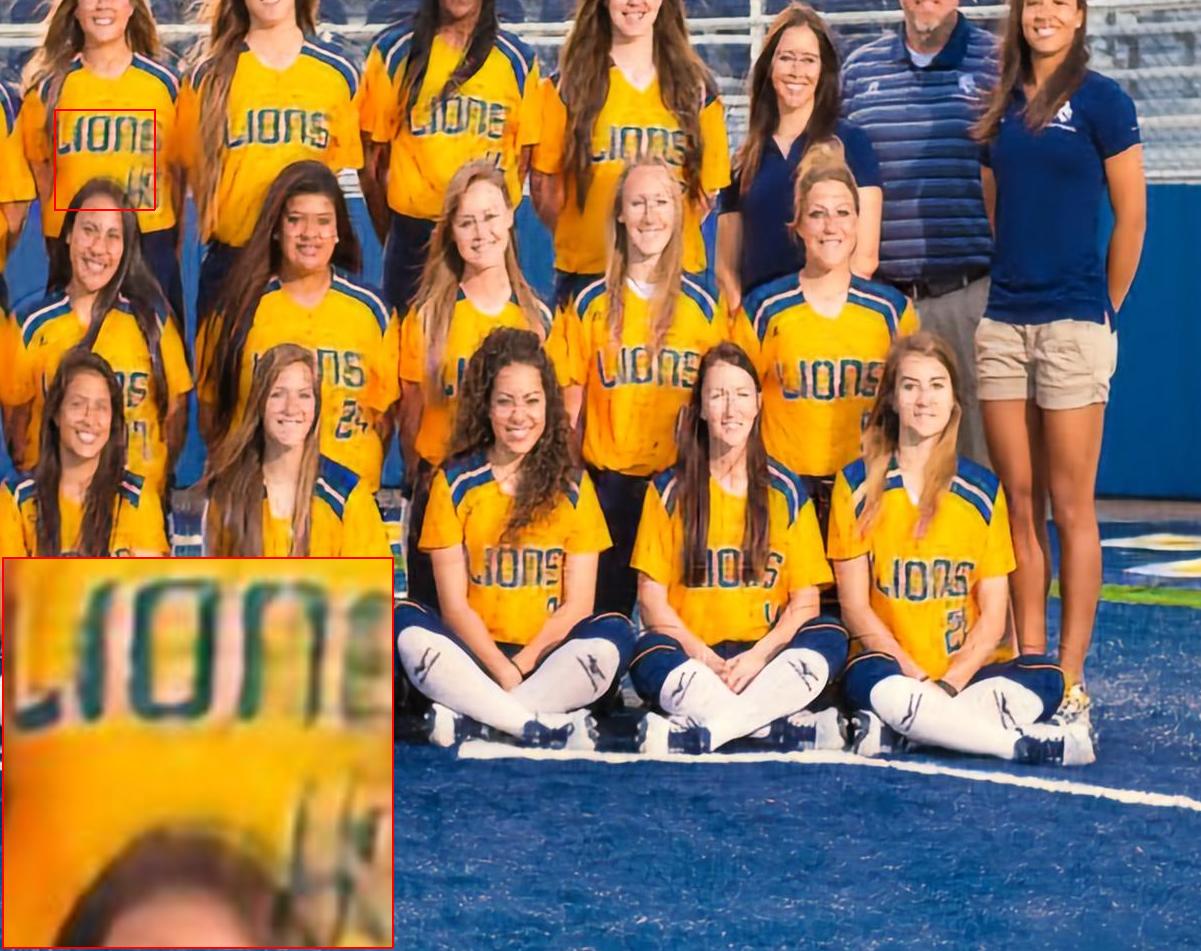}
		\subcaption*{MIMO-UNet}
	\end{subfigure}
	\centering
	\begin{subfigure}{0.24\linewidth}
	\setlength{\abovecaptionskip}{0.cm}
		\centering
		\includegraphics[width=\linewidth]{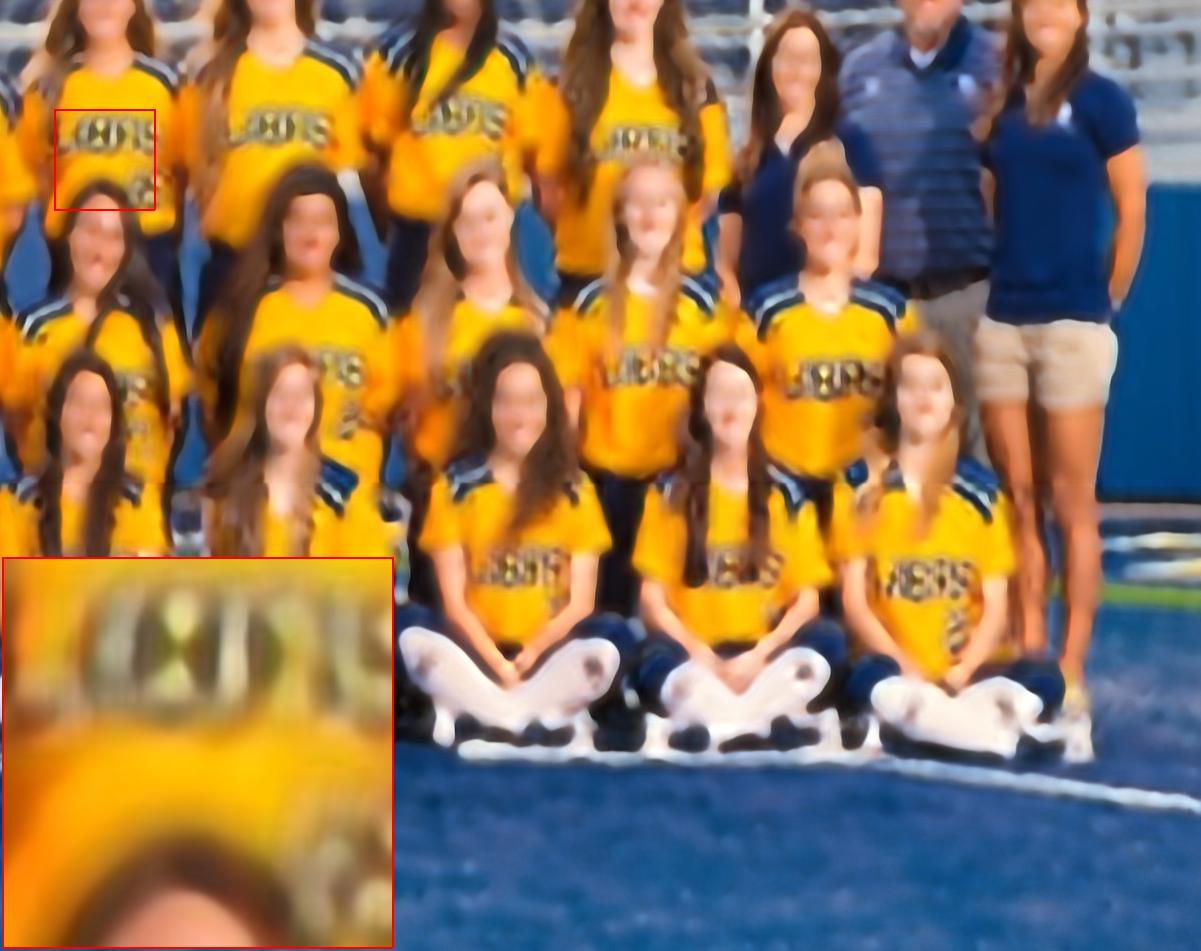}
		\subcaption*{MPRNet}
	\end{subfigure}
	\centering
	\begin{subfigure}{0.24\linewidth}
	\setlength{\abovecaptionskip}{0.cm}
		\centering
		\includegraphics[width=\linewidth]{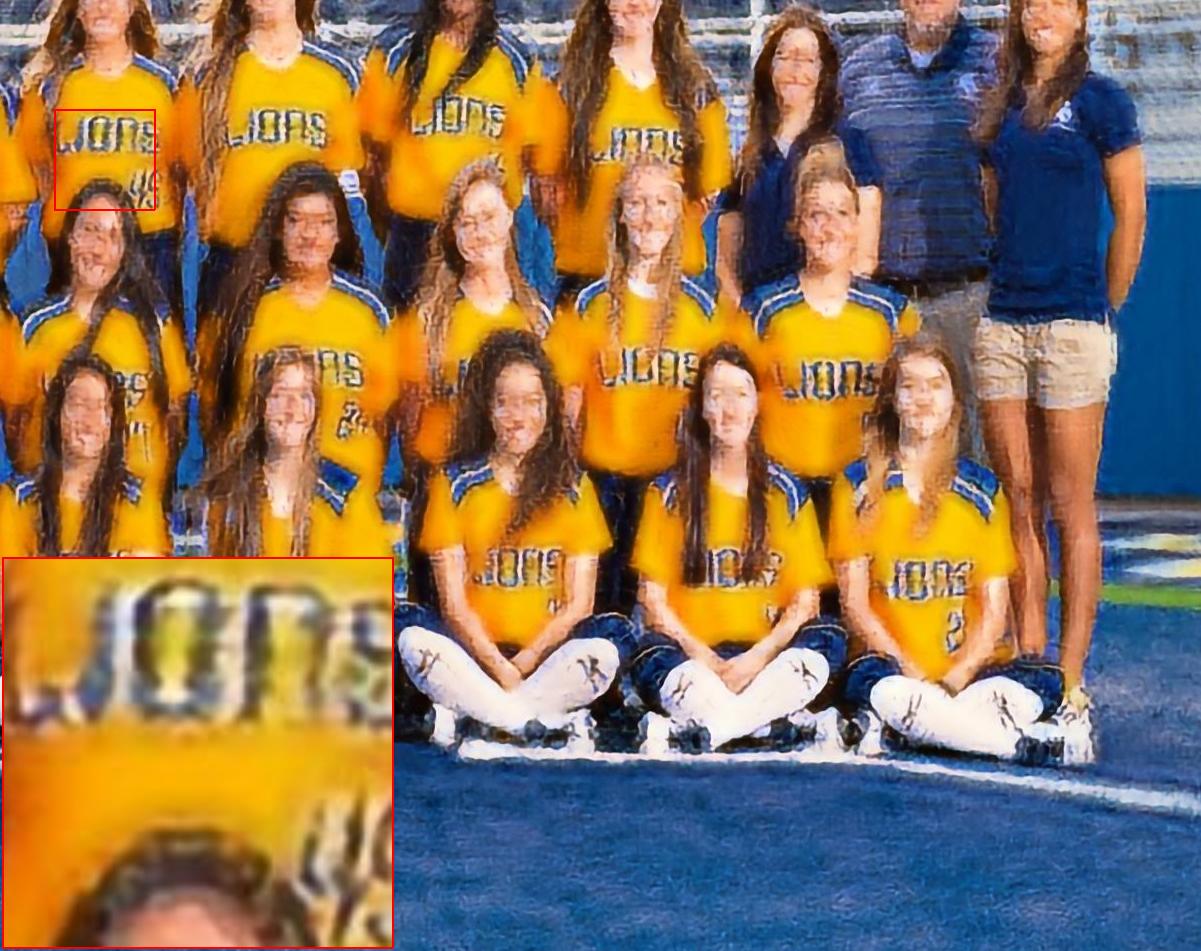}
		\subcaption*{Stripformer}
	\end{subfigure}
	\centering
	\begin{subfigure}{0.24\linewidth}
	\setlength{\abovecaptionskip}{0.cm}
		\centering
		\includegraphics[width=\linewidth]{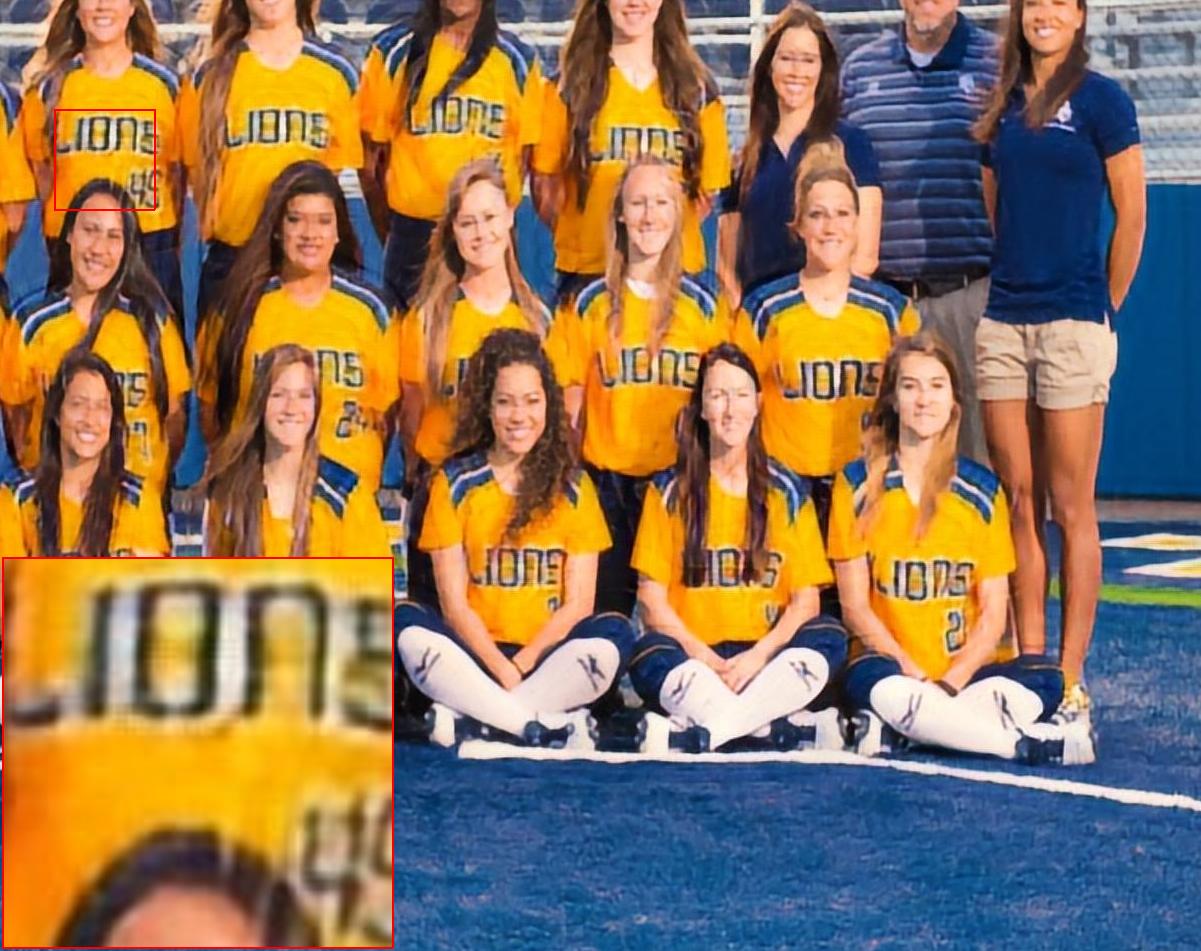}
		\subcaption*{Ours}
	\end{subfigure}
	\centering
	\begin{subfigure}{0.24\linewidth}
	\setlength{\abovecaptionskip}{0.cm}
		\centering
		\includegraphics[width=\linewidth]{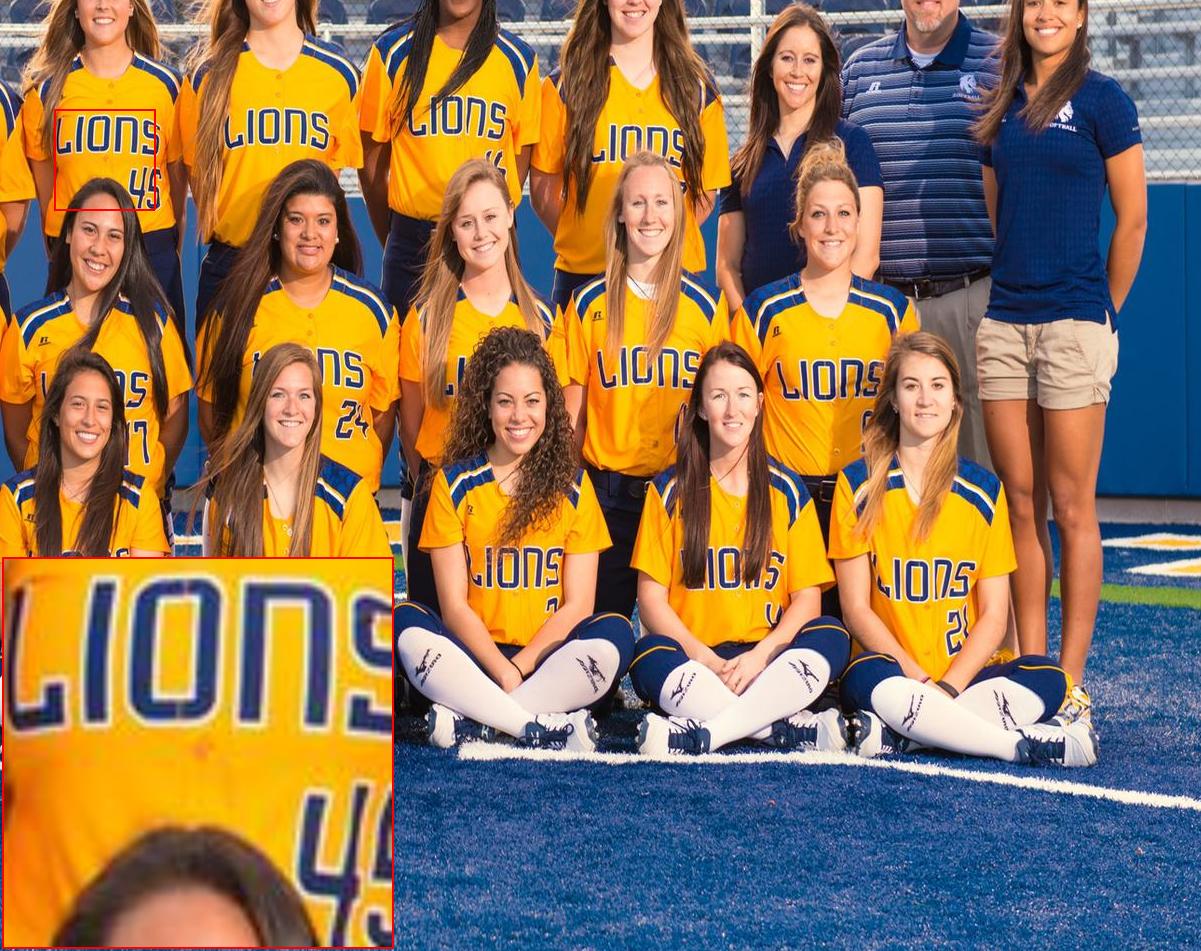}
		\subcaption*{Reference}
	\end{subfigure}
	\caption{\textbf{Qualitative comparisons on the synthetic dataset.} The results are produced by DeblurGANv2\cite{kupyn2019deblurgan}, FOV-KPN\cite{chen2021extreme}, MIMO-UNet\cite{cho2021rethinking}, MPRNet\cite{zamir2021multi}, Stripformer\cite{tsai2022stripformer} and our method. Here, "Distance" represents the imaging object distance to the focal plane. The first scene is obtained by cropping the image patch from the ISO 12233 test chart. The second and third scenes are from DIV2K dataset \cite{agustsson2017ntire}.}
	\label{synthetic}
\end{figure}

\subsection{Evaluation on Synthetic Images}
\label{sec:4.2}
To demonstrate the advantages of the proposed method, we firstly compare it with five state-of-the-art methods on the synthetic test images, including DeblurGANv2 \cite{kupyn2019deblurgan}, FOV-KPN \cite{chen2021extreme}, MIMO-UNet \cite{cho2021rethinking}, MPRNet \cite{zamir2021multi}, Stripformer \cite{tsai2022stripformer}. For a fair comparison, all the compared methods adopt the default settings from the original papers. The training/test dataset consists of 4000/1000 degradation images, which are generated by the proposed imaging simulation framework and include 101 degradation degrees. The compared methods are retrained on the synthetic training dataset. We evaluate these methods through commonly used metrics, including Peak Signal Noise Ratio (PSNR), Structural Similarity Index (SSIM) \cite{wang2004image}, Learned Perceptual Image Patch Similarity (LPIPS) \cite{zhang2018unreasonable}.

Table \ref{synthetic-table} and Fig. \ref{synthetic} show the quantitative and qualitative results of our method and the compared methods on the synthetic test dataset, respectively. Our method outperforms the previous best method MIMO-UNet by 0.34dB in PSNR, as shown in Table \ref{synthetic-table}. Fig. \ref{synthetic} demonstrates that our method can recover images with better quality than all the compared methods. Our method yields sharper results with more details, especially in highly textured regions such as the text on the clothes. Furthermore, our method benefits from the invertible design and the combined forward loss and reverse loss, which effectively avoids unrealistic information and artifacts in the results.

\begin{figure}[h!]
	\centering
	\begin{subfigure}{0.24\linewidth}
	\setlength{\abovecaptionskip}{0.cm}
	\setlength{\belowcaptionskip}{0.2cm}
		\centering
		\includegraphics[width=\linewidth]{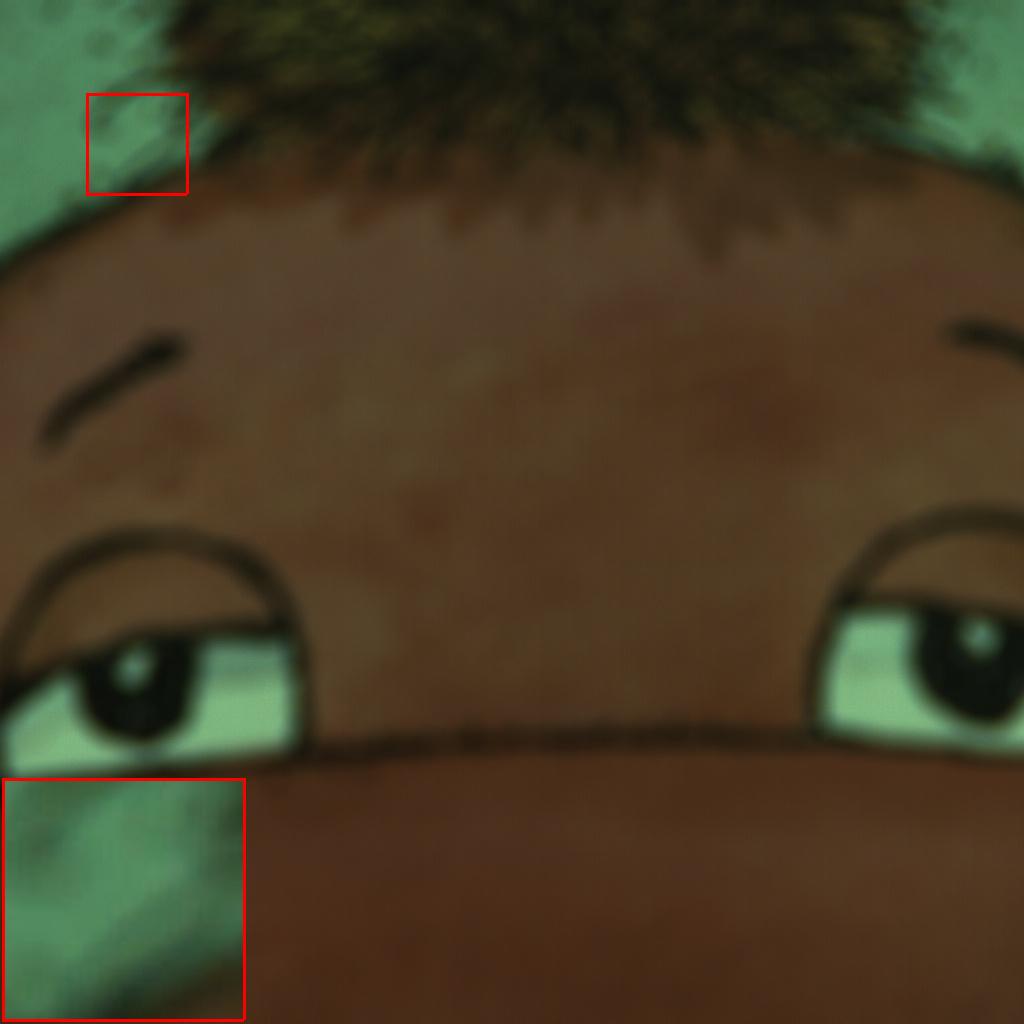}
		\subcaption*{Input(Distance:123mm)}
	\end{subfigure}
	\centering
	\begin{subfigure}{0.24\linewidth}
	\setlength{\abovecaptionskip}{0.cm}
	\setlength{\belowcaptionskip}{0.2cm}
		\centering
		\includegraphics[width=\linewidth]{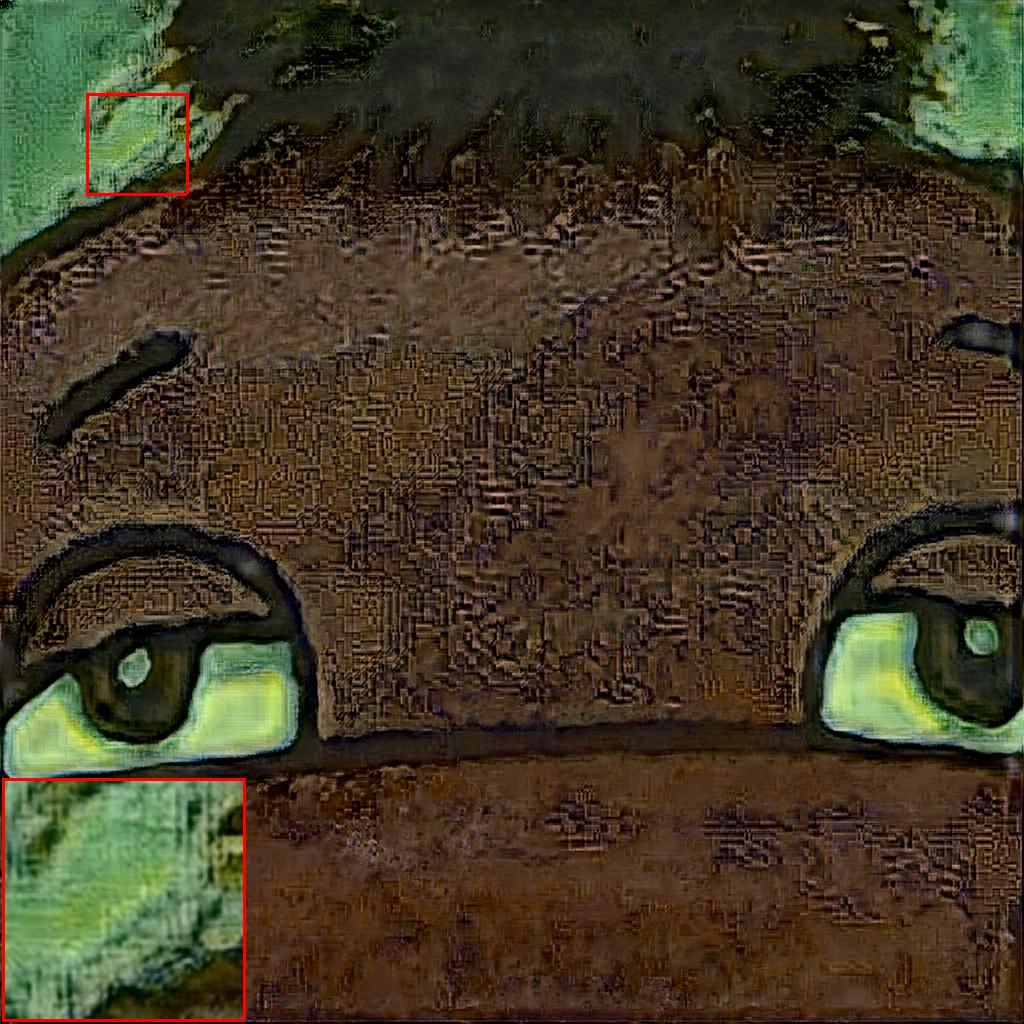}
		\subcaption*{DeblurGANv2}
	\end{subfigure}
	\centering
	\begin{subfigure}{0.24\linewidth}
	\setlength{\abovecaptionskip}{0.cm}
	\setlength{\belowcaptionskip}{0.2cm}
		\centering
		\includegraphics[width=\linewidth]{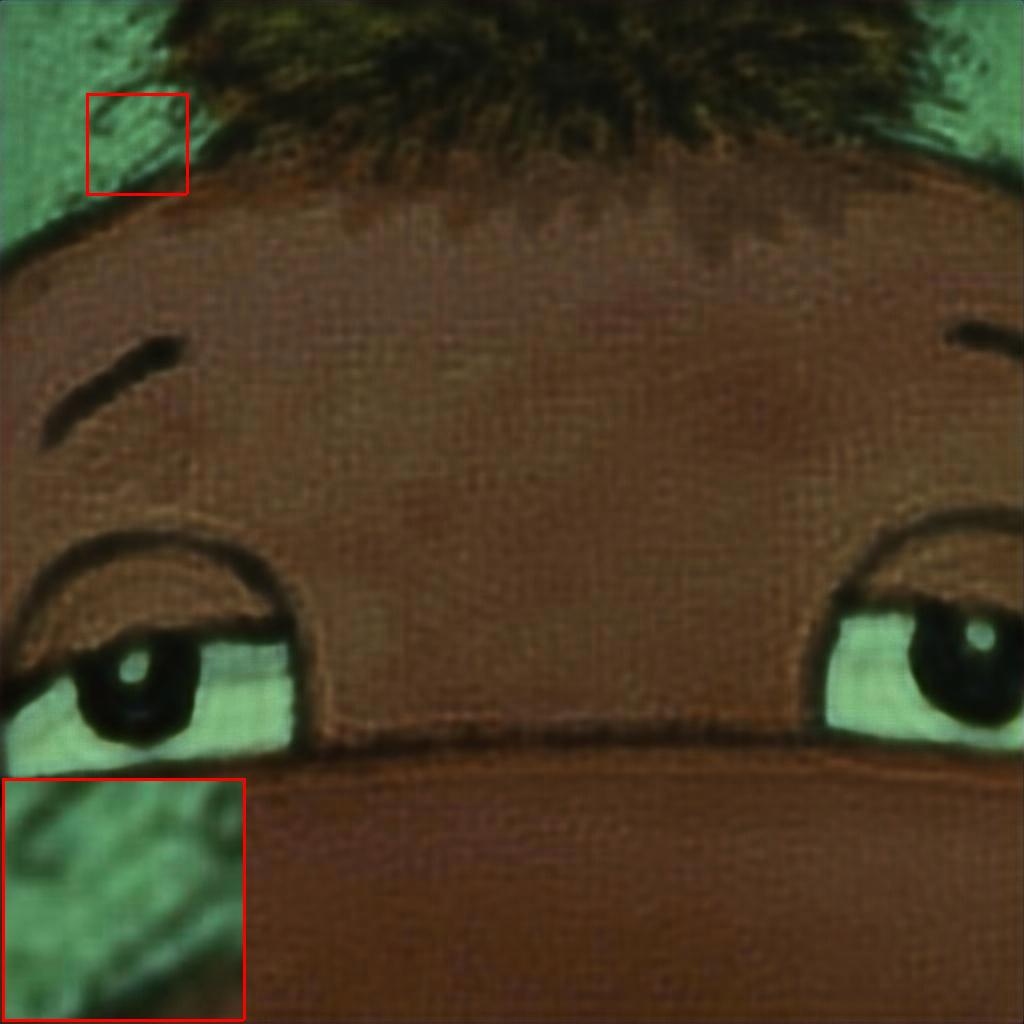}
		\subcaption*{FOV-KPN}
	\end{subfigure}
	\centering
	\begin{subfigure}{0.24\linewidth}
	\setlength{\abovecaptionskip}{0.cm}
	\setlength{\belowcaptionskip}{0.2cm}
		\centering
		\includegraphics[width=\linewidth]{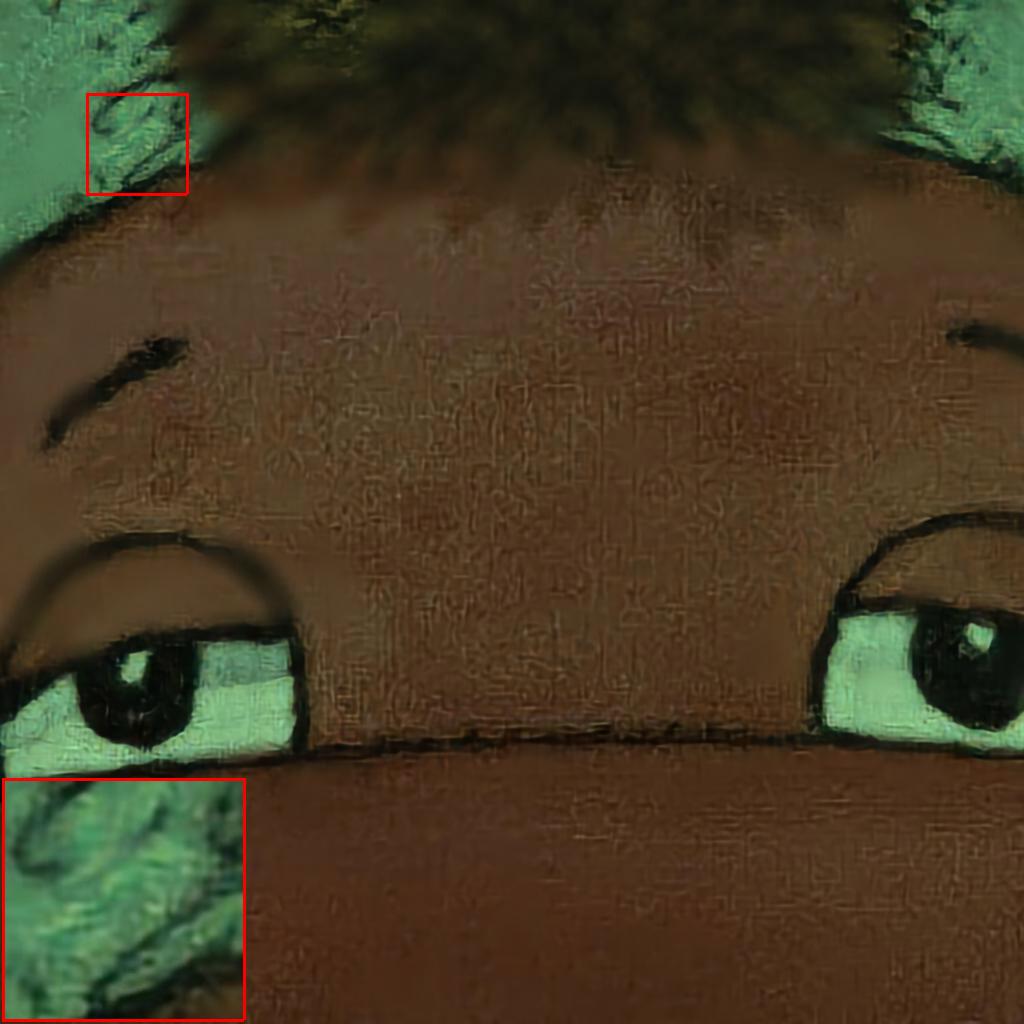}
		\subcaption*{MIMO-UNet}
	\end{subfigure}
	\centering
	\begin{subfigure}{0.24\linewidth}
	\setlength{\abovecaptionskip}{0.cm}
	\setlength{\belowcaptionskip}{0.2cm}
		\centering
		\includegraphics[width=\linewidth]{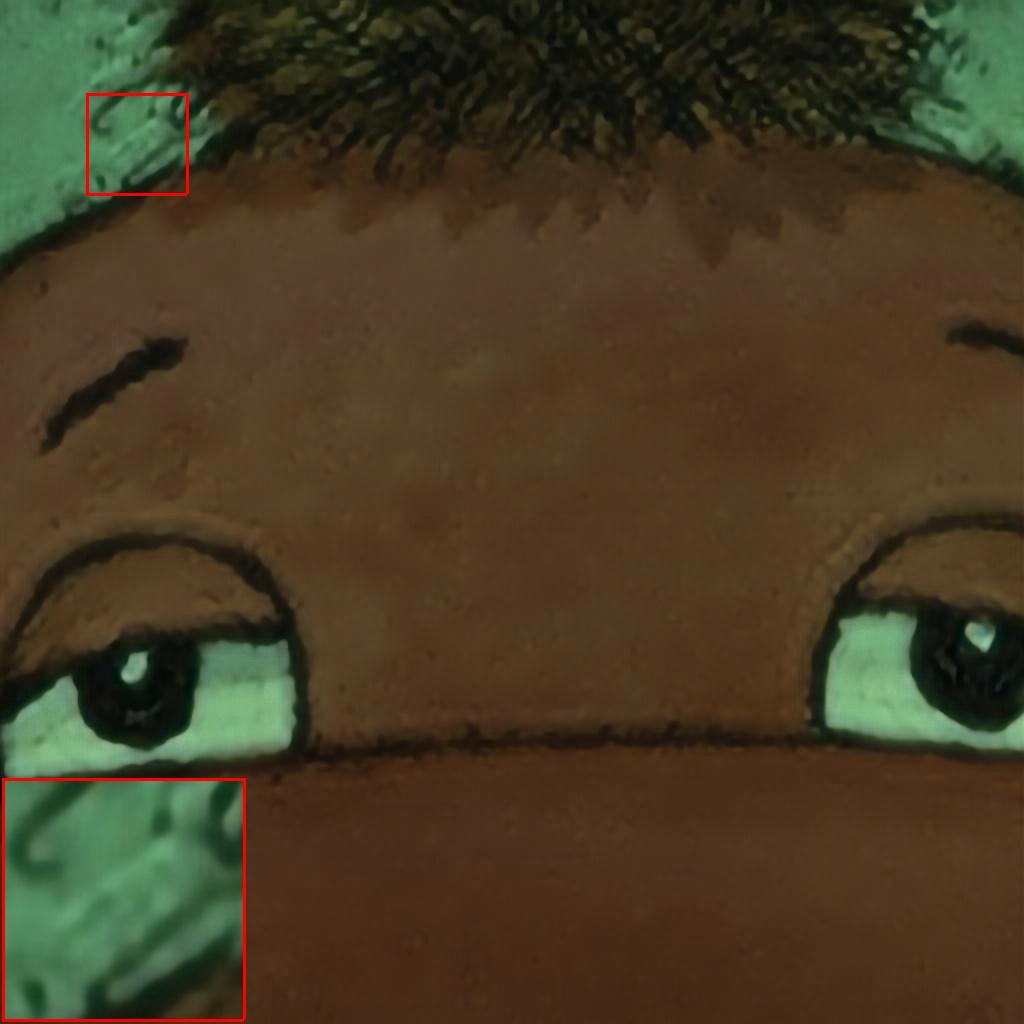}
		\subcaption*{MPRNet}
	\end{subfigure}
	\centering
	\begin{subfigure}{0.24\linewidth}
	\setlength{\abovecaptionskip}{0.cm}
	\setlength{\belowcaptionskip}{0.2cm}
		\centering
		\includegraphics[width=\linewidth]{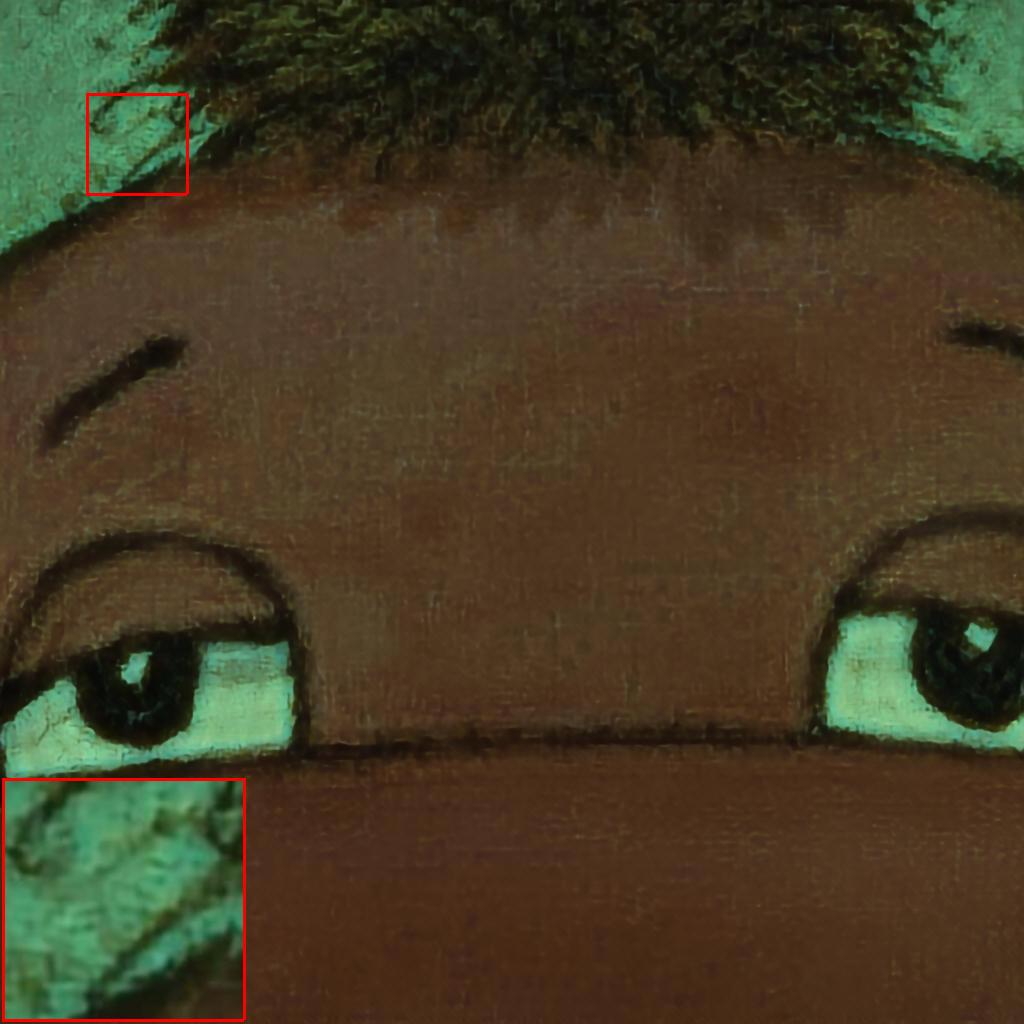}
		\subcaption*{Stripformer}
	\end{subfigure}
	\centering
	\begin{subfigure}{0.24\linewidth}
	\setlength{\abovecaptionskip}{0.cm}
	\setlength{\belowcaptionskip}{0.2cm}
		\centering
		\includegraphics[width=\linewidth]{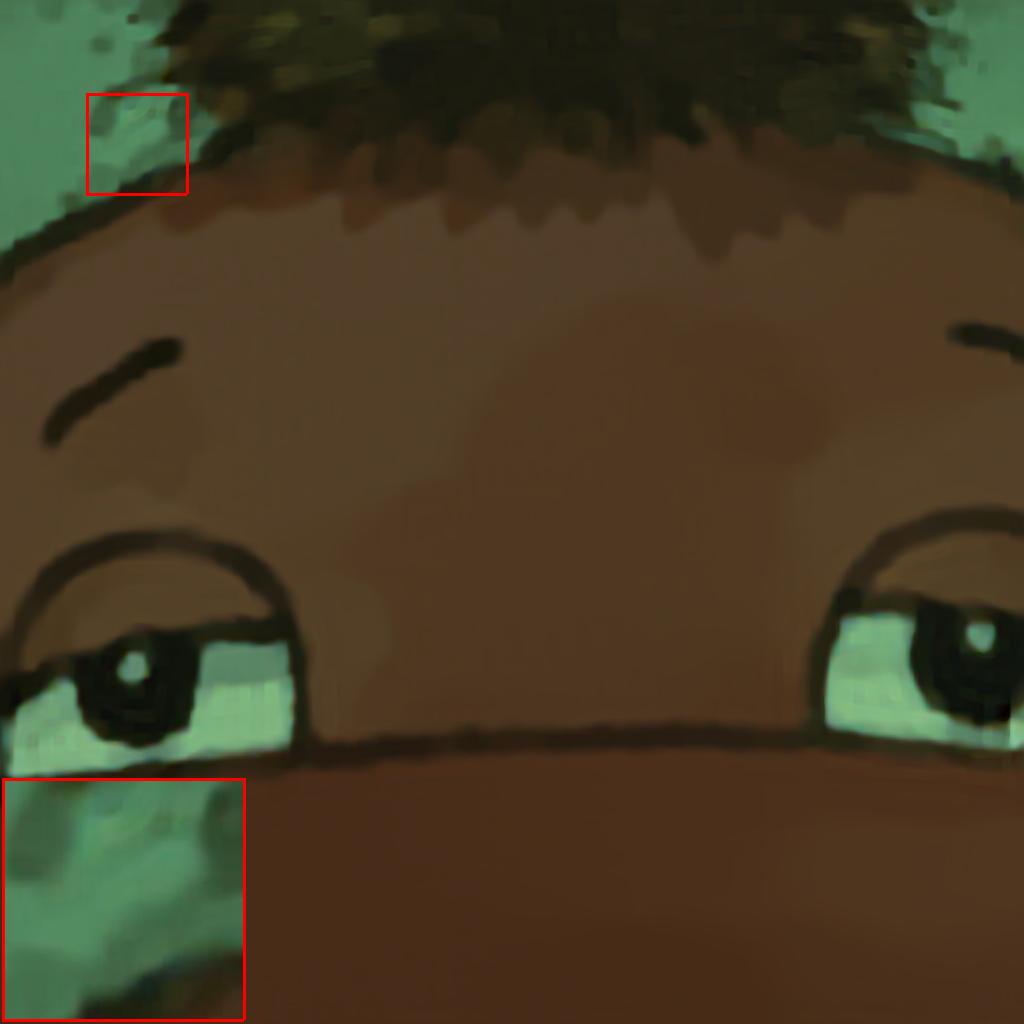}
		\subcaption*{Dark Channel Prior}
	\end{subfigure}
	\centering
	\begin{subfigure}{0.24\linewidth}
	\setlength{\abovecaptionskip}{0.cm}
	\setlength{\belowcaptionskip}{0.2cm}
		\centering
		\includegraphics[width=\linewidth]{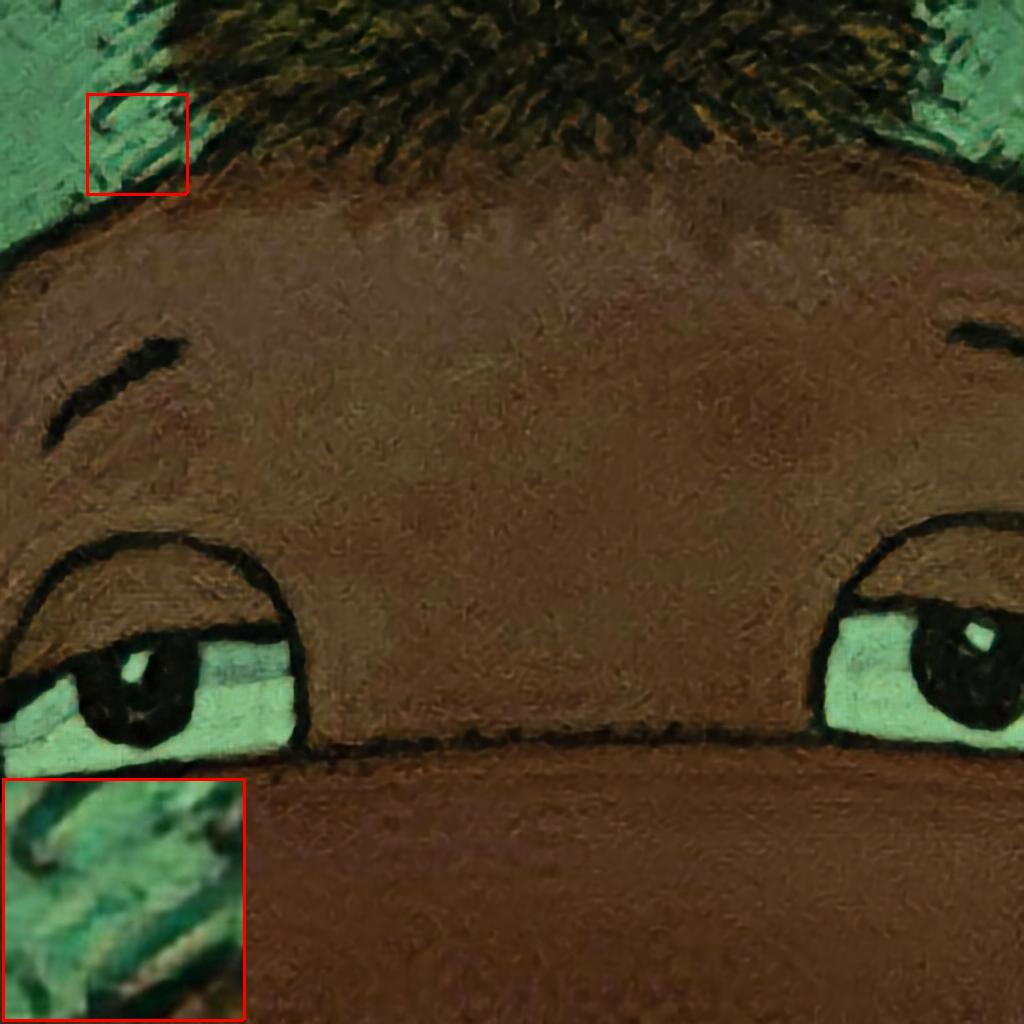}
		\subcaption*{Ours}
	\end{subfigure}
	
	\centering
	\begin{subfigure}{0.24\linewidth}
	\setlength{\abovecaptionskip}{0.cm}
	\setlength{\belowcaptionskip}{0.2cm}
		\centering
		\includegraphics[width=\linewidth]{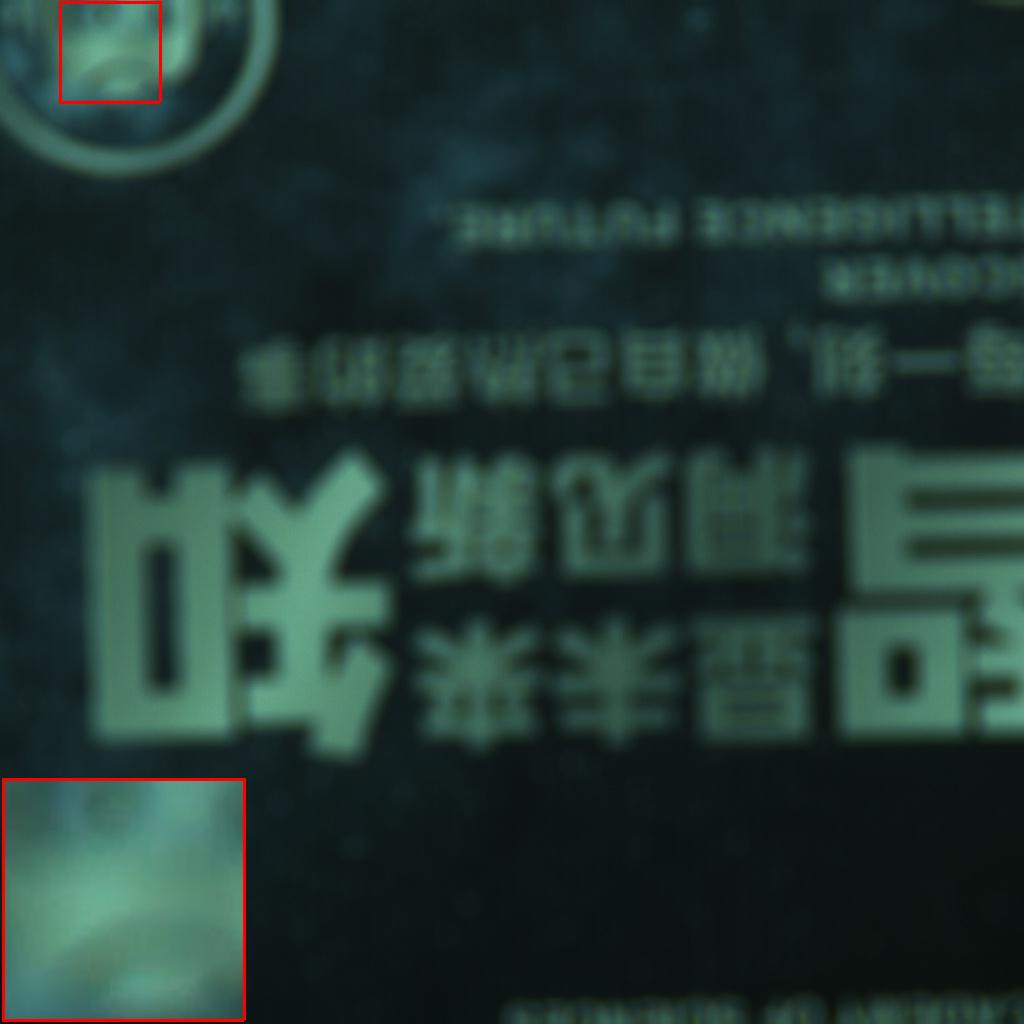}
		\subcaption*{Input(Distance:110mm)}
	\end{subfigure}
	\centering
	\begin{subfigure}{0.24\linewidth}
	\setlength{\abovecaptionskip}{0.cm}
	\setlength{\belowcaptionskip}{0.2cm}
		\centering
		\includegraphics[width=\linewidth]{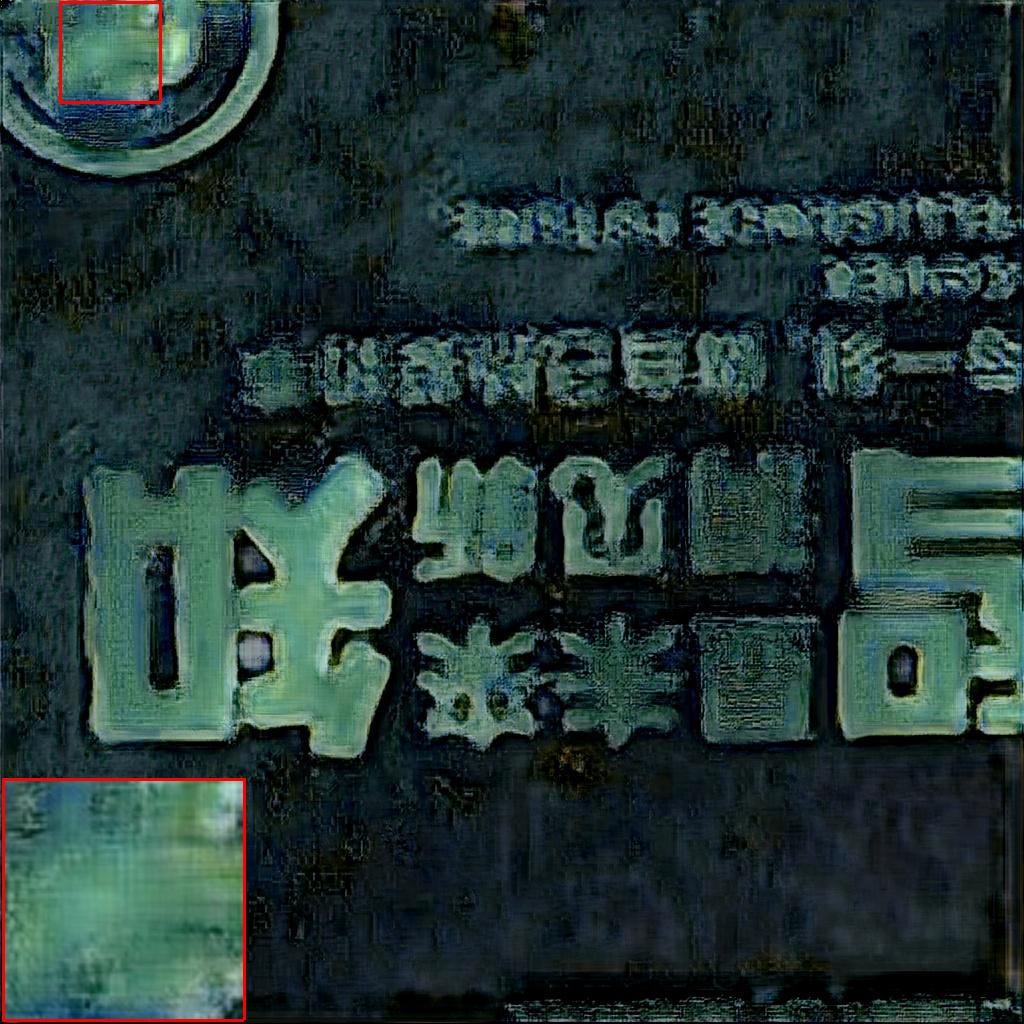}
		\subcaption*{DeblurGANv2}
	\end{subfigure}
	\centering
	\begin{subfigure}{0.24\linewidth}
	\setlength{\abovecaptionskip}{0.cm}
	\setlength{\belowcaptionskip}{0.2cm}
		\centering
		\includegraphics[width=\linewidth]{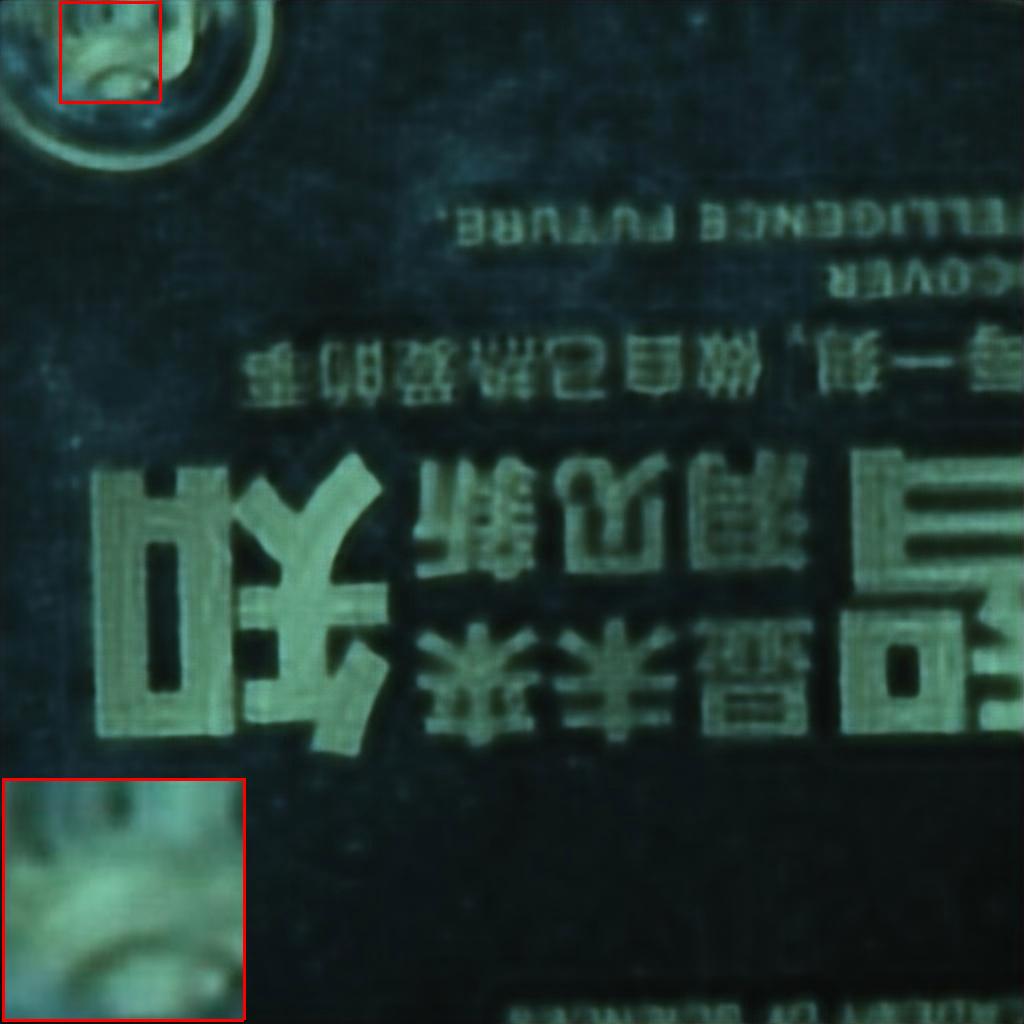}
		\subcaption*{FOV-KPN}
	\end{subfigure}
	\centering
	\begin{subfigure}{0.24\linewidth}
	\setlength{\abovecaptionskip}{0.cm}
	\setlength{\belowcaptionskip}{0.2cm}
		\centering
		\includegraphics[width=\linewidth]{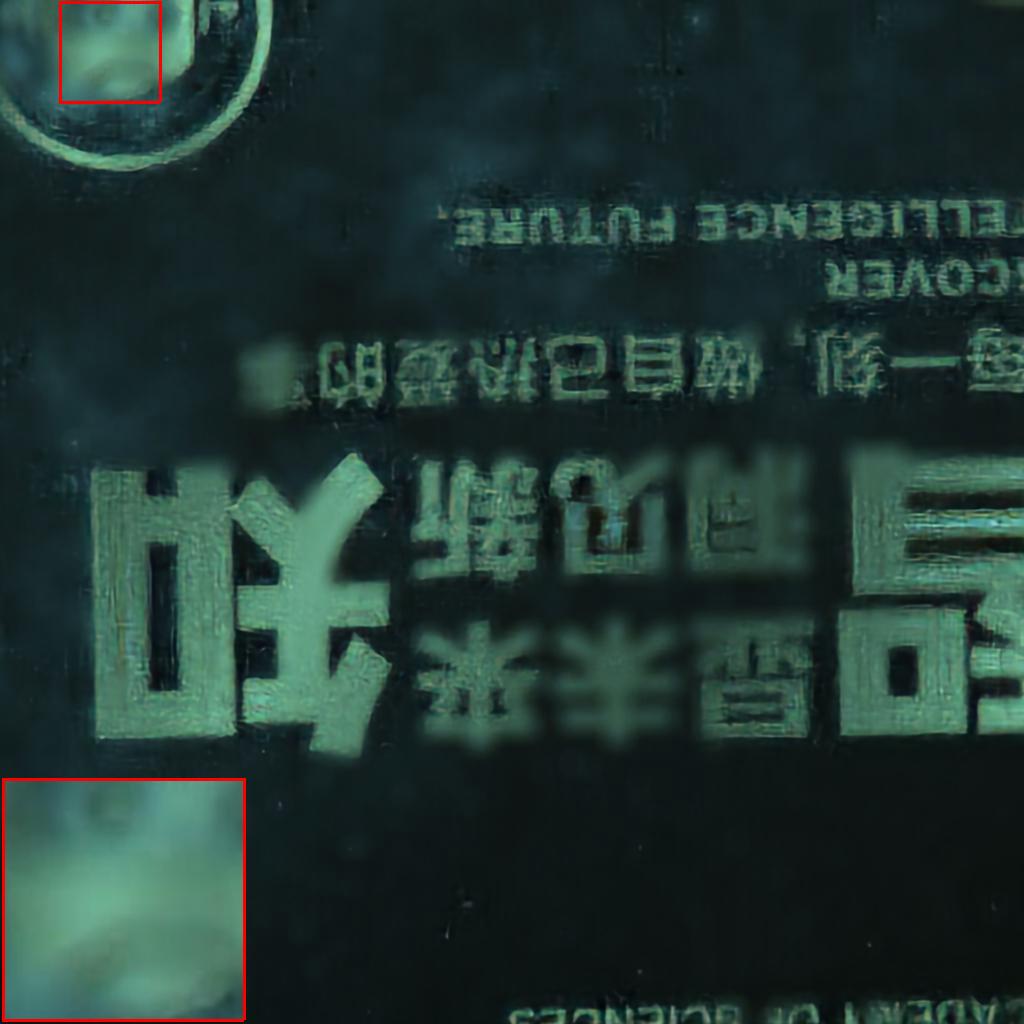}
		\subcaption*{MIMO-UNet}
	\end{subfigure}
	\centering
	\begin{subfigure}{0.24\linewidth}
	\setlength{\abovecaptionskip}{0.cm}
		\centering
		\includegraphics[width=\linewidth]{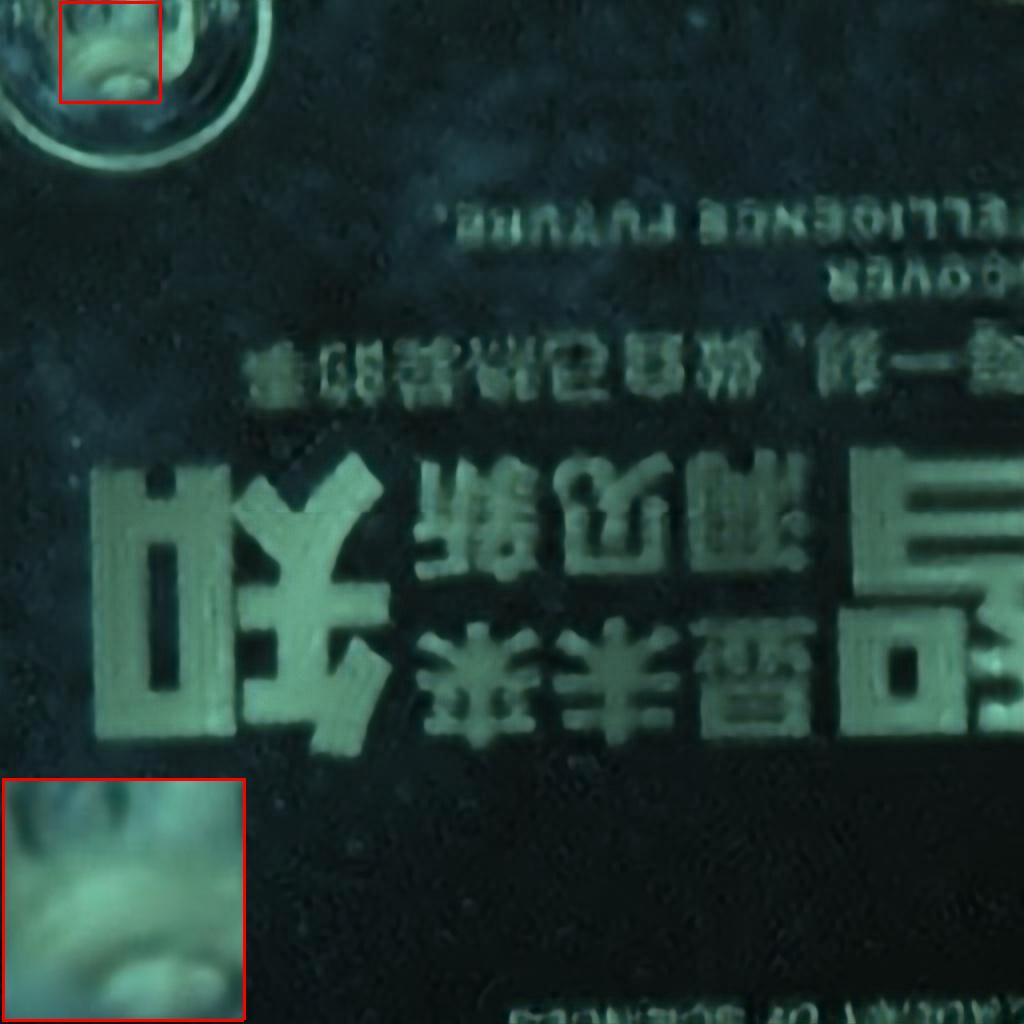}
		\subcaption*{MPRNet}
	\end{subfigure}
	\centering
	\begin{subfigure}{0.24\linewidth}
	\setlength{\abovecaptionskip}{0.cm}
		\centering
		\includegraphics[width=\linewidth]{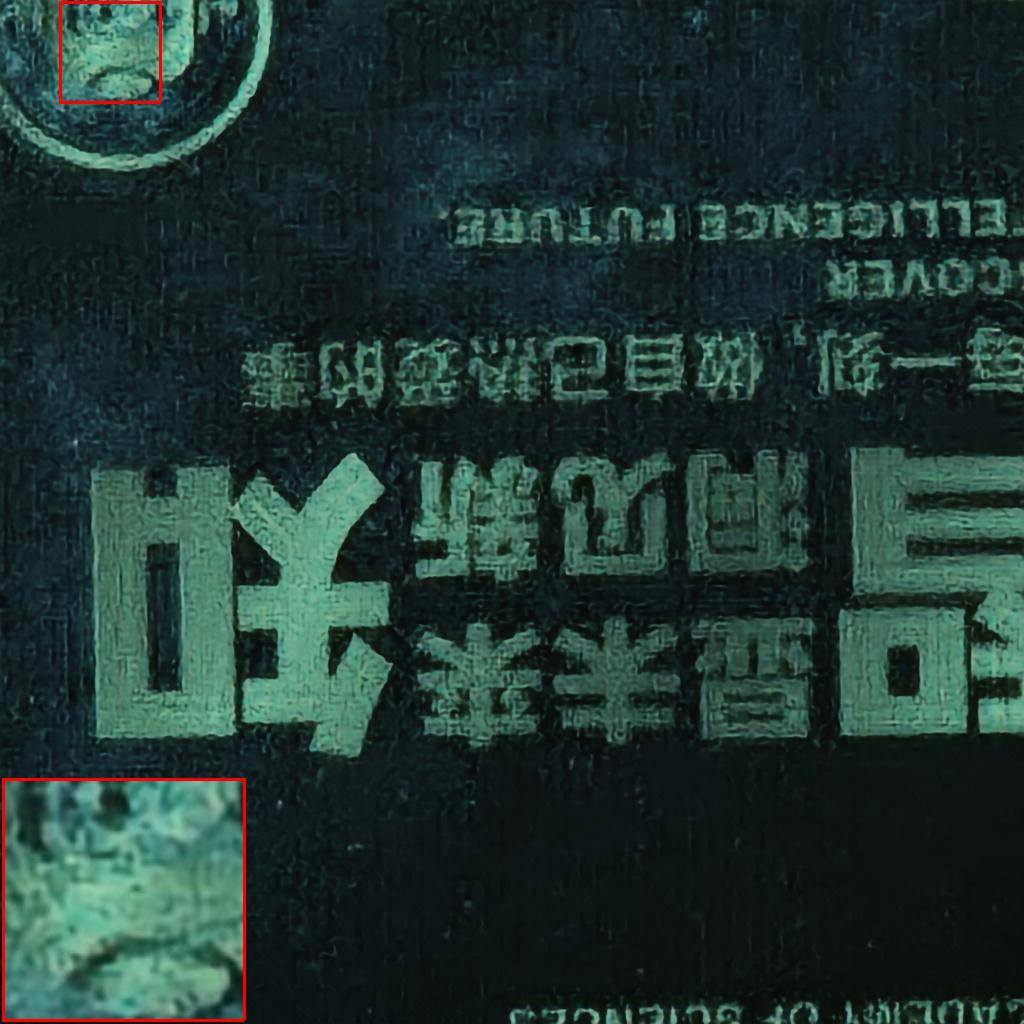}
		\subcaption*{Stripformer}
	\end{subfigure}
	\centering
	\begin{subfigure}{0.24\linewidth}
	\setlength{\abovecaptionskip}{0.cm}
		\centering
		\includegraphics[width=\linewidth]{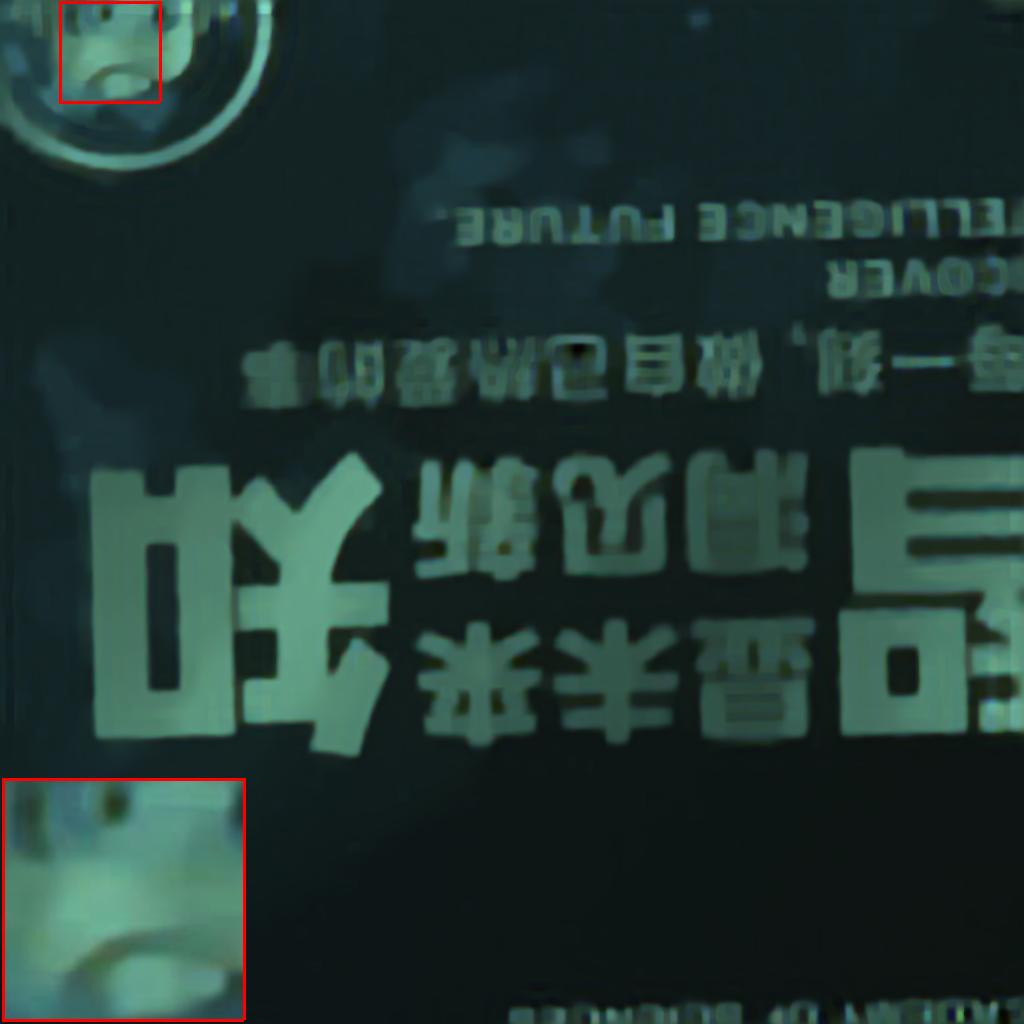}
		\subcaption*{Dark Channel Prior}
	\end{subfigure}
	\centering
	\begin{subfigure}{0.24\linewidth}
	\setlength{\abovecaptionskip}{0.cm}
		\centering
		\includegraphics[width=\linewidth]{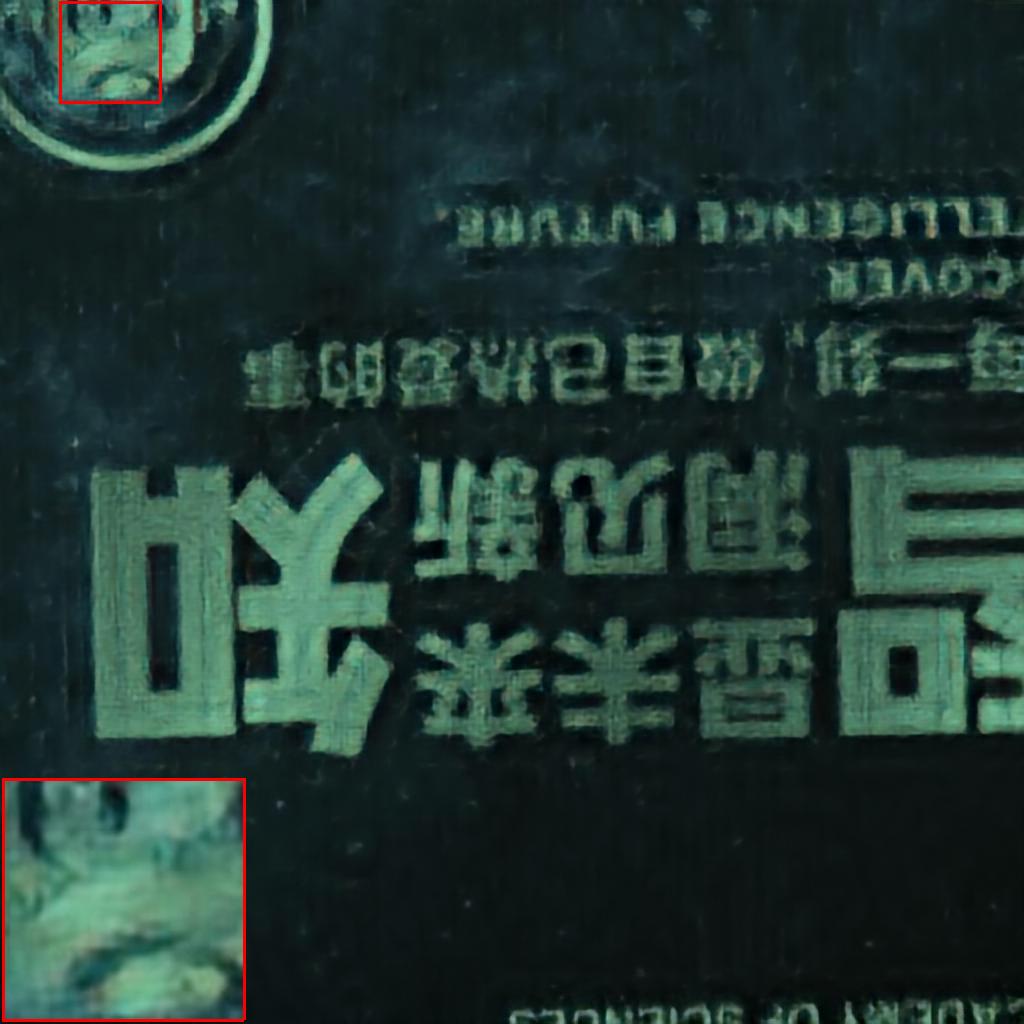}
		\subcaption*{Ours}
	\end{subfigure}
	\caption{\textbf{Qualitative comparisons on the real test images.} The restored results from
left to right are produced by DeblurGANv2\cite{kupyn2019deblurgan}, FOV-KPN\cite{chen2021extreme}, MIMO-UNet\cite{cho2021rethinking}, MPRNet\cite{zamir2021multi}, Stripformer\cite{tsai2022stripformer}, Dark Channel Prior\cite{pan2016blind} and our method. Here, "Distance" represents the imaging object distance to the focal plane.}
    \label{real-image}
\end{figure}

\subsection{Evaluation on Real Images}
In Fig. \ref{real-image}, we compare our method with the state-of-the-art methods on realistic test images captured by our experimental camera. In addition to the methods mentioned in the Section \ref{sec:4.2}, we add the model-driven optimization algorithm \cite{pan2016blind} for comparison. For convenience, we refer to \cite{pan2016blind} as “Dark Channel Prior”. As can be seen from Fig. \ref{real-image}, the proposed method outperforms all methods in terms of the visual effect. It effectively eliminates the degradation caused by optical aberrations, and the conditional invertible neural network can largely retain details of the original image, such as text edges and hair structures. Although the DeblurGANv2 recovers images with sharper edges, it introduced unrealistic information and severe noise. It is worth noting that MIMO-UNet\cite{cho2021rethinking}, which ranks second in quantitative results on the synthetic dataset, cannot handle severe spatially variant degradation. Other methods fail to deal with the optical aberrations well, and the restored images are not sufficiently clear. Overall, our method performs better in resolving optical aberrations on real images.

\begin{figure}[htbp]
    \centering
    \includegraphics[width=0.8\linewidth]{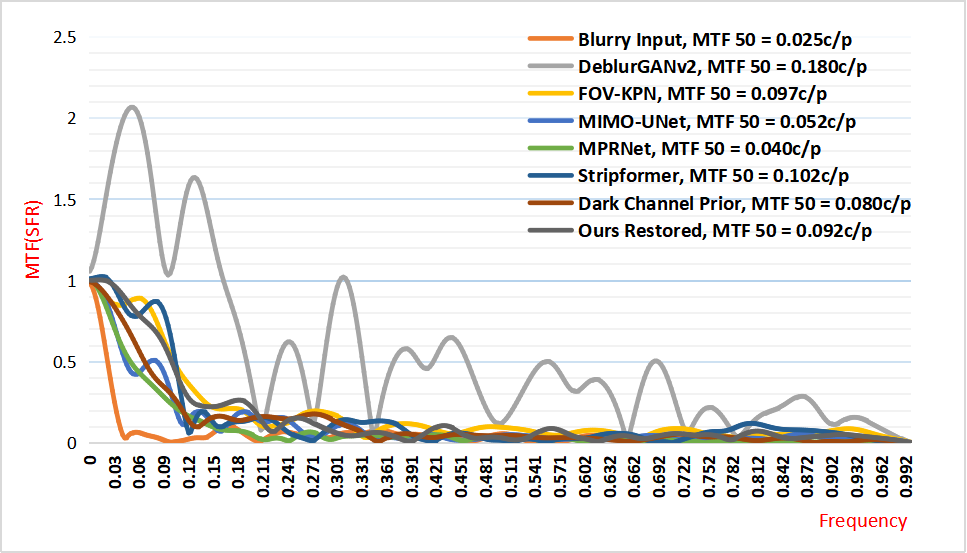}
	\caption{The MTF curve of the real captured image and the corresponding restored sharp image produced by our method and other models.}
	\label{mtf}
\end{figure}

To further evaluate the improvement of image quality, we analyze the MTF of restored images. The second scene of Fig. \ref{real-image} is used because there are many edges of text. We show the MTF curves of the degraded input, the sharp image restored by our methods, and the images restored by other methods for comparison, which can be seen in Fig. \ref{mtf}. Our method improves the MTF50 from 0.025 to 0.092 c/p, demonstrating that our method generates images with sharp edges. Although DeblurGANv2 produces a higher MTF50 value, their model leads to over-sharp edges which are unrealistic and causes severe noise which can not be reflected by the MTF curve. The restored image by Stripformer \cite{tsai2022stripformer} also suffers from severe noise which can be seen in the blue background.

\subsection{Ablation Studies}
In this section, we evaluate the impact of every component of our method by ablating different parts of the neural network and comparing them with the complete architecture, as shown in Table \ref{ablation-table}. We conduct ablation experiments on the synthetic dataset. Details about the dataset can be found in Section \ref{sec:4.1}. We use the Adam \cite{kingma2015adam} optimizer with a learning rate of 0.0001 to train for 150 epochs, and the learning rate decreases by half every 50 epochs.

\begin{table}
\begin{center}
\caption{Quantitative results of ablation studies in terms of PSNR, SSIM and LPIPS on test dataset. The feature extraction module (FEM), the conditional code (CC), the invertible module (IM) and the FOV encoder (FOV).}
\setlength{\tabcolsep}{4mm}{
\begin{tabular}{cccc|ccc}
\hline FEM & CC & IN & FOV & PSNR $\uparrow$ & SSIM $\uparrow$ & LPIPS$\downarrow$\\
\hline & $\sqrt{ }$ & $\sqrt{ }$ & & 23.79 & 0.7989 & 0.3107\\
 $\sqrt{ }$ & & $\sqrt{ }$ & & 23.85 & 0.8042 & 0.2751\\
 $\sqrt{ }$ & & & & 23.21 & 0.7963 & 0.2628\\
 $\sqrt{ }$ & $\sqrt{ }$ & $\sqrt{ }$ & $\sqrt{ }$ & 24.22 & 0.8170 &0.2530\\
 $\sqrt{ }$ & $\sqrt{ }$ & $\sqrt{ }$ & & $\mathbf{25.30}$ & $\mathbf{0.8432}$ & $\mathbf{0.2069}$ \\
\hline
\end{tabular}}
\label{ablation-table}
\end{center}
\end{table}

\begin{figure}[h!]
	\centering
	\begin{subfigure}{0.24\linewidth}
		\centering
		\includegraphics[width=\linewidth]{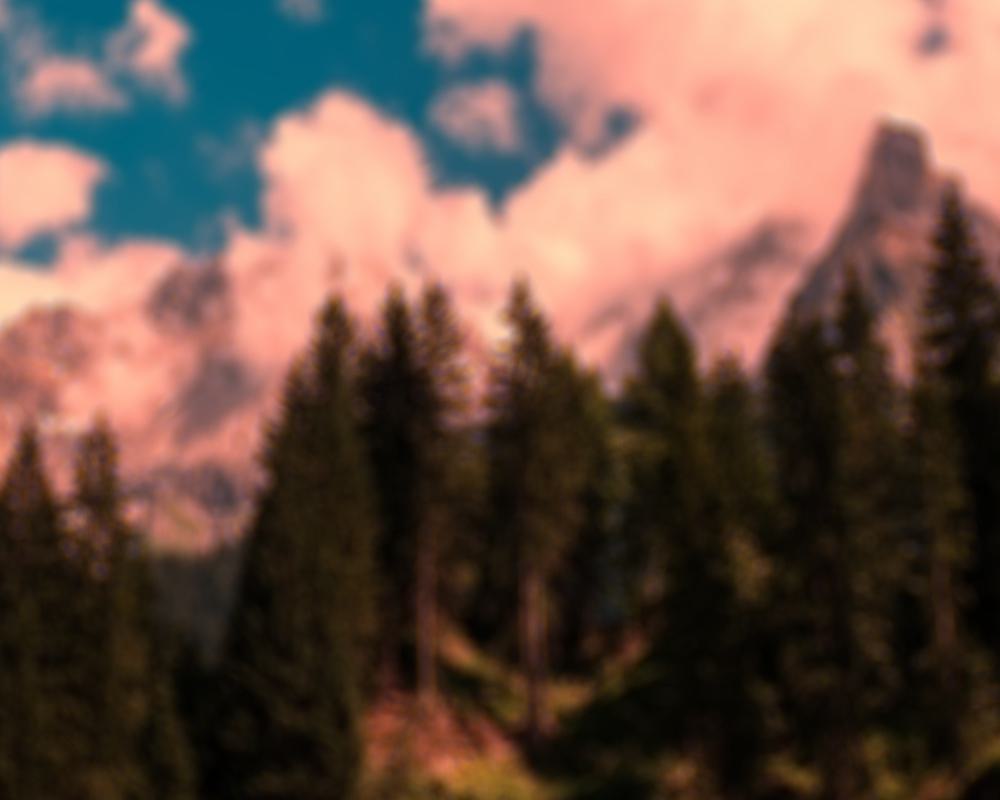}
		\subcaption*{Blurred Input}
	\end{subfigure}
	\centering
	\begin{subfigure}{0.24\linewidth}
		\centering
		\includegraphics[width=\linewidth]{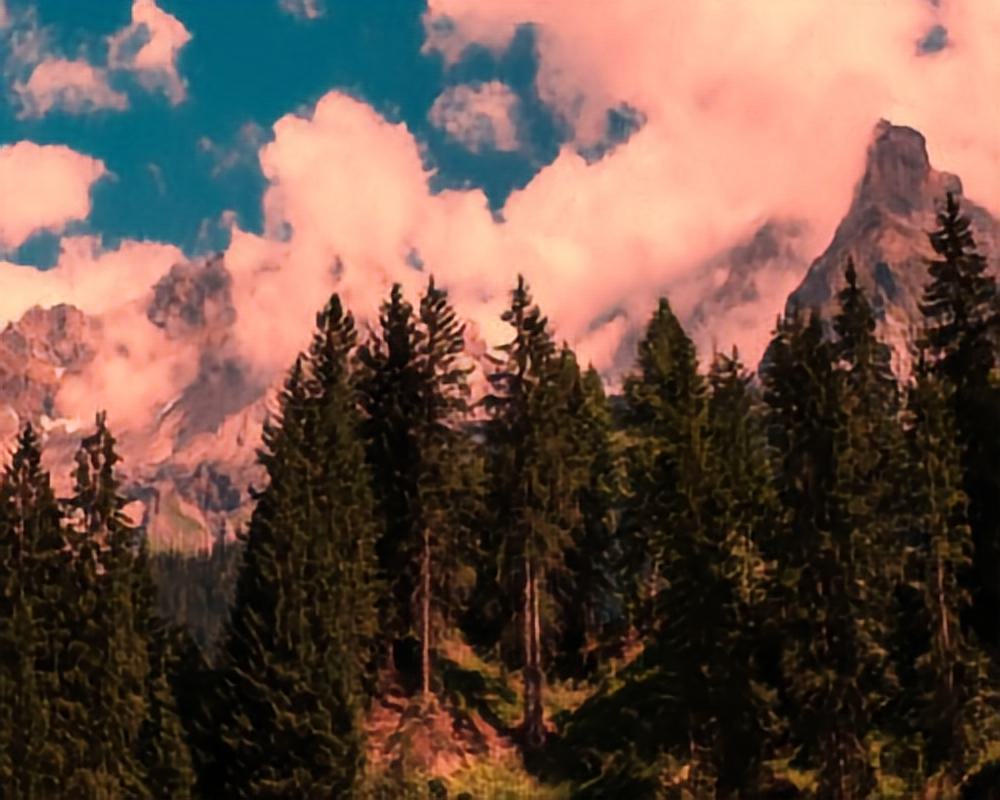}
		\subcaption*{Ours}
	\end{subfigure}
	\centering
	\begin{subfigure}{0.24\linewidth}
		\centering
		\includegraphics[width=\linewidth]{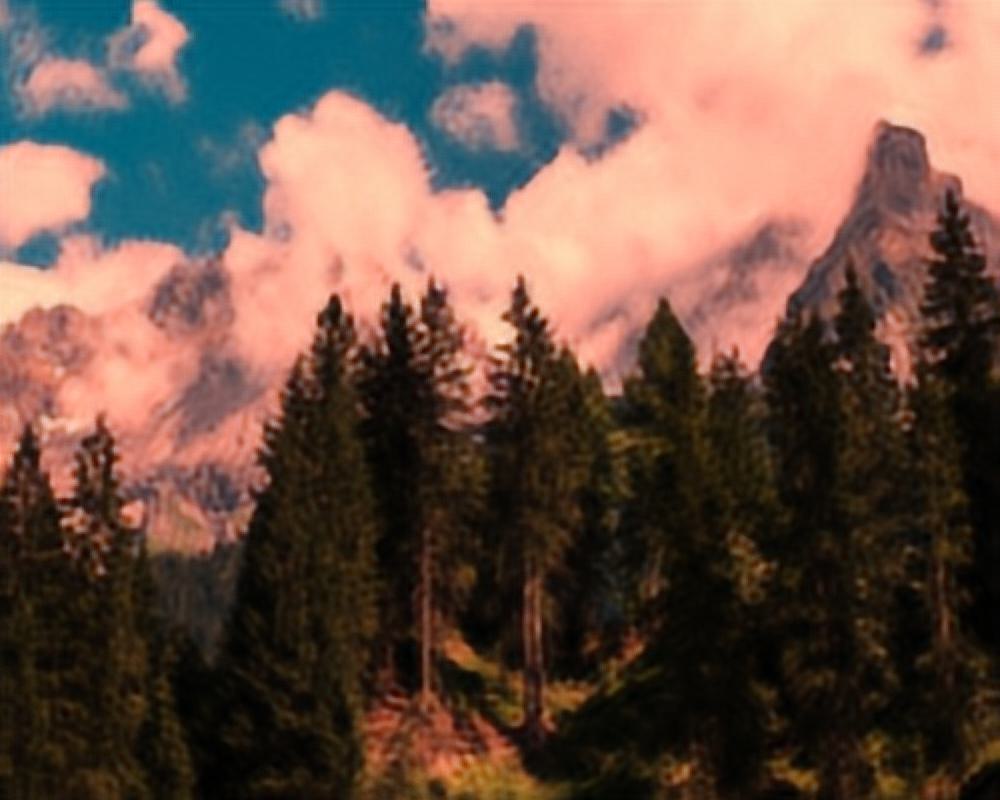}
		\subcaption*{Without Feature Extraction}
	\end{subfigure}
	\centering
	\begin{subfigure}{0.24\linewidth}
		\centering
		\includegraphics[width=\linewidth]{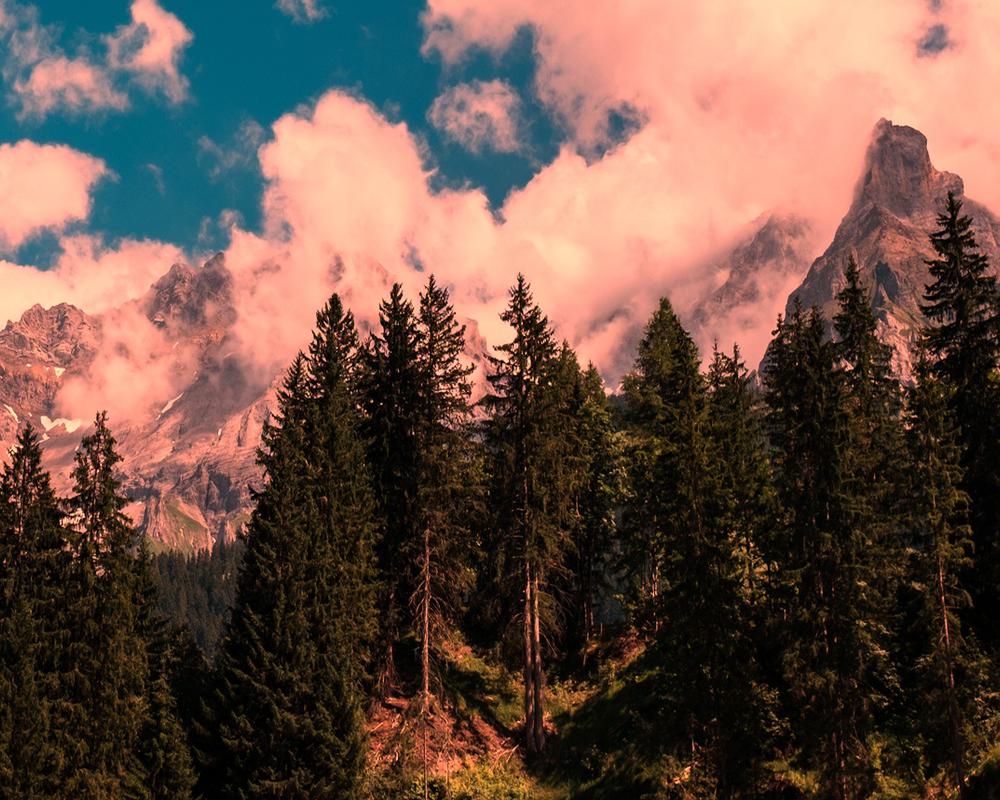}
		\subcaption*{Reference}
	\end{subfigure}
	\caption{\textbf{Ablation study on feature extraction module.} This image is from DIV2K dataset\cite{agustsson2017ntire}.}
    \label{without-feature extraction}
\end{figure}

\textbf{Analysis of the feature extraction module.} To verify the performance of the proposed nonlinear feature extraction module, we conduct a corresponding ablation study on the module, as shown in Table \ref{ablation-table}. Specifically, we train the proposed method with and without the feature extraction module respectively and keep the other training settings the same. Fig. \ref{without-feature extraction} shows the visualizations of the two models evaluated on the test dataset, and we can observe that using the feature enhancement module effectively improves the image quality.

\textbf{Analysis of the conditional invertible blocks.} Removing the conditional code, the quantitative scores significantly drop as shown in the second row of Table \ref{ablation-table}. It means that the conditional code is crucial for image fidelity and perceptual quality. After removing the entire conditional invertible module and only keeping the feature extraction module as shown in the third row, the aberrations correcting ability of the network gets worse, which demonstrates the necessity of the conditional invertible module.

\textbf{Analysis of the FOV encoder \cite{chen2021extreme}.} The work in \cite{chen2021optical} found that adding the field of view (FOV) as an additional input can improve the model performance. We also try to add a FOV encoder \cite{chen2021extreme} in front of the forward process. However, the model performance is lower than the proposed method without FOV as can be seen in Table \ref{ablation-table}. Thus, we do not include the FOV encoder in the proposed method.

\begin{table}
\arrayrulecolor{black}
\begin{center}
\caption{Comparisons on the effect of different numbers of conditional invertible blocks.}
\begin{tabular}{c|ccc|c|c}
\hline Conditional Invertible Blocks & PSNR $\uparrow$ & SSIM $\uparrow$ & LPIPS$\downarrow$ & Params$\#$ & Time\\
\hline k=8 & 25.16 & 0.8387 & 0.2105 & 7.67M & 0.739s\\
 k=12 & 25.30 & 0.8432 & 0.2069 & 10.16M & 0.921s\\
 k=16 & 25.74 & 0.8447 & 0.1978 & 12.65M & 1.162s\\
\hline
\end{tabular}
\label{block-number}
\end{center}
\end{table}

\textbf{Analysis of the number of conditional invertible blocks.} We verify the effect of different numbers of conditionally reversible blocks on the optical aberration correction performance of our method in Table \ref{block-number}. Reducing the number $k$ of conditionally reversible blocks will lead to artifacts in the restored results, and the PSNR and SSIM also decrease significantly, such as when $k=8$. When increasing the number of blocks to 12, the proposed method is able to recover more image details, and the performance of the model is greatly improved. When the number of blocks is further increased to 16, the performance improvement of the model is not obvious. However, the increase in the number of reversible blocks means an increase in the number of parameters and inference time, so in order to better balance model performance and efficiency, we use $k=12$ as the default option.

\begin{table}
\begin{center}
\caption{Comparisons on the effect of the proposed loss functions.}
\begin{tabular}{cccc|ccc}
\hline $\mathcal{L}_{forward}$ &  $\mathcal{L}_{reverse}$ & $\mathcal{L}_{edge}$ & $\mathcal{L}_{perceptual}$ &  PSNR $\uparrow$ & SSIM $\uparrow$ & LPIPS$\downarrow$\\
\hline $\sqrt{ }$ & $\sqrt{ }$ & $\sqrt{ }$ & $\sqrt{ }$ & $\mathbf{25.30}$ & $\mathbf{0.8432}$ & $\mathbf{0.2069}$\\
$\sqrt{ }$ &  & $\sqrt{ }$  & $\sqrt{ }$ & 24.89 & 0.8312 & 0.2310 \\
$\sqrt{ }$ & $\sqrt{ }$ & & $\sqrt{ }$& 23.91 & 0.8020 &  0.2755\\
$\sqrt{ }$ & $\sqrt{ }$ & $\sqrt{ }$ & &23.63 & 0.7954 & 0.3126\\
\hline
\end{tabular}
\label{loss-table}
\end{center}
\end{table}

\textbf{Analysis of the proposed loss functions.}
The loss functions are applied during the training stage to minimize the difference between the restored image and the ground truth image, as described in section \ref{sec:3.3}. We conduct ablation studies on different loss functions to verify the impact of them, and the results are shown in Table \ref{loss-table}. We noticed that the proposed method without any of the reverse loss, edge loss, and perceptual loss leads to worse performance. The reverse loss makes the model more stable. Edge loss and perceptual loss minimize the difference between two images from the perspective of high-frequency details and perceptual quality, enabling the model to generate sharper images.

\section{Conclusions}
In this work, we have proposed an enhanced conditional invertible neural framework to correct variable-degree optical aberrations. The conditional invertible neural network can effectively avoid information loss and restore image details. Meanwhile, to better handle different degrees of aberrations, we embed the degree of degradation into the model as a conditional encoding. Comprehensive experiments verify that our method outperforms compared methods in correcting optical aberrations on both synthetic images and real images. Furthermore, our method performs quite competitively in terms of model size. The proposed method is promising to be embedded into ISP systems to improve imaging quality.

\begin{backmatter}
\bmsection{Funding}
Ministry of Science and Technology of the People's Republic of China (2021YFB3601404); Chinese Academy of Sciences (E2RC5901).

\bmsection{Disclosures}
The authors declare no conflicts of interest.

\bmsection{Data availability}
Data underlying the results presented in this paper are not publicly available at this time but may be obtained from the authors upon reasonable request.

\end{backmatter}


\end{document}